\documentclass{article}

\usepackage[main, final]{neurips_2026}
\usepackage[utf8]{inputenc} % allow utf-8 input
\usepackage[T1]{fontenc}    % use 8-bit T1 fonts
\usepackage{hyperref}       % hyperlinks
\usepackage{url}            % simple URL typesetting
\usepackage{booktabs}       % professional-quality tables
\usepackage{nicefrac}       % compact symbols for 1/2, etc.
\usepackage{microtype}      % microtypography
\usepackage{xcolor}         % colors

\usepackage{amsmath}
\usepackage{amssymb}
\usepackage{mathtools}
\usepackage{amsthm}

\usepackage{bm}
\usepackage{multirow}
\usepackage{color, colortbl}
\usepackage{algorithm}
\usepackage{listings}
\usepackage{subcaption}
\usepackage{threeparttable}
\usepackage{enumitem}
\usepackage{hhline}
\usepackage{wrapfig}
\usepackage{needspace}

\theoremstyle{plain}
\newtheorem{theorem}{Theorem}[section]

\theoremstyle{definition}

\theoremstyle{remark}

\newcommand{\xx}{$\bm{x}$}
\newcommand{\ee}{$\bm{\varepsilon}$}
\newcommand{\vv}{$\bm{v}$}

\newcommand{\diff}{\mathrm{d}}
\newcommand{\Diff}{\mathcal{D}}
\newcommand{\DIFF}{\widetilde{\mathcal{D}}}
\newcommand{\E}[1]{\mathbb{E}\left[#1\right]}

\newcommand{\norm}[1]{\lVert#1\rVert}
\DeclareMathOperator{\Tr}{Tr}

\newlength\savewidth\newcommand\shline{\noalign{\global\savewidth\arrayrulewidth
  \global\arrayrulewidth 1pt}\hline\noalign{\global\arrayrulewidth\savewidth}}

\newcommand\midline{\noalign{\global\savewidth\arrayrulewidth
  \global\arrayrulewidth 0.5pt}\hline\noalign{\global\arrayrulewidth\savewidth}}

\newcommand{\tablestyle}[2]{\setlength{\tabcolsep}{#1}\renewcommand{\arraystretch}{#2}\centering\footnotesize}

\definecolor{LightOrange}{rgb}{1.0,0.9,0.8}
\definecolor{LightPurple}{rgb}{1.0,0.8,0.9}
\definecolor{LightCyan}{rgb}{0.6,0.8,1.0}

\newcommand{\hhs}{\hspace{-0.001em}}
\newcommand{\vvs}{\vspace{-.1em}}

\newlength{\plotwidth}
\newcolumntype{x}[1]{>{\centering\arraybackslash}p{#1pt}}
\newcolumntype{y}[1]{>{\raggedright\arraybackslash}p{#1pt}}
\newcolumntype{z}[1]{>{\raggedleft\arraybackslash}p{#1pt}}

\title{Understanding and Learning Prediction Targets in Diffusion Models}

\author{%
  Qing Jin\thanks{Project lead} \\
  Independent Researcher \\
  \texttt{jinqingking@gmail.com} \\
  \And
  Chaoyang Wang \\
  Independent Researcher \\
  \texttt{} \\
}

\begin{document}

\maketitle
\begin{abstract}
% This document provides a basic paper template and submission guidelines.
% Abstracts must be a single paragraph, ideally between 4--6 sentences long.
% Gross violations will trigger corrections at the camera-ready phase.
Recent advances in diffusion and flow matching models have highlighted a shift in the preferred prediction target---moving from noise (\ee) and velocity (\vv) to direct data (\xx) prediction---particularly in high-dimensional settings.
% However, a formal understanding of why the optimal target varies across different tasks remains elusive.
However, a formal explanation of why the optimal target depends on the specific properties of the data remains elusive.
% In this work, we provide a theoretical framework based on a generalized prediction formulation that accommodates arbitrary output targets, of which \ee-, \vv-, and \xx-prediction are special cases.
In this work, we provide a theoretical framework based on a generalized prediction formulation $\bm{u}=k\bm{x}-(1-k)\bm{n}$, where $\bm{x}$ represents clean data and $\bm{n}$ denotes noise.
This formulation accommodates arbitrary targets, of which~\ee-,~\vv-, and~\xx-prediction are special cases.
We derive the analytical relationship between data's geometry and the optimal prediction target, offering a rigorous justification for why \xx-prediction becomes superior when the ambient dimension significantly exceeds the data's intrinsic dimension.
% In addition, while our theory identifies dimensionality as the underlying factor to determine the optimal prediction, the intrinsic dimension of manifold-bound data is typically intractable to estimate in practice.
Furthermore, our controlled analysis reveals how time conditioning and loss variations systematically modulate this optimal objective.
While our theory identifies dimensionality as the governing factor for the optimal prediction target, the intrinsic dimension of manifold-bound data is typically intractable to estimate in practice.
% In addition, since the intrinsic dimension of manifold-bound data is typically intractable to estimate in practice,
% To bridge this gap, we propose $\bm{k}$-\textbf{Diff}, a framework that employs a data-driven method to learn the optimal prediction parameter $k$ without requiring explicit dimension information.
% To bridge this gap, we propose $\bm{k}$\textbf{-Diff}, a framework that employs a data-driven approach to learn the optimal prediction parameter $k$ directly from data, bypassing the need for explicit dimension estimation.
% To bridge this gap, we propose $\bm{k}$\textbf{-Diff}, a framework that employs a data-driven approach to learn the optimal prediction parameter $k$ directly from data. We define the output prediction as $\bm{u} = k\bm{x} - (1-k)\bm{n}$, where $\bm{x}$ represents the clean data and $\bm{n}$ denotes the noise.
To bridge this gap, we propose $\bm{k}$\textbf{-Diff}, a framework that learns the optimal parameter $k$ directly from data, bypassing the need for explicit dimension estimation.
% intractable intrinsic dimension estimation.
Extensive experiments in both latent-space and pixel-space image generation demonstrate that $k$-Diff consistently outperforms fixed-target baselines across varying architectures and data scales, providing a principled and automated approach to enhancing generative performance.
\end{abstract}
\section{Introduction}
% Diffusion probabilistic models and flow matching frameworks have revolutionized generative modeling, enabling high-fidelity synthesis across domains ranging from image generation to video and 3D modeling.
% A central design choice in these frameworks is the selection of the prediction target.
% Early seminal works primarily utilized noise prediction ($\varepsilon$-prediction); however, this was largely superseded by velocity prediction ($v$-prediction).
% The shift toward $v$-prediction was driven by its improved training stability, straighter sampling paths, and the ability to achieve high-quality results with significantly fewer sampling steps.
% Most recently, research into high-resolution generation---particularly for models operating directly in pixel space---has revealed that data prediction ($x$-prediction) often yields superior results and enhanced stability compared to other options.

One of the key design choices of diffusion models~\cite{sohl2015deep} is the network output prediction.
Early works on low-dimensional data primarily utilize noise prediction (\ee-prediction)~\cite{ho2020denoising,song2020score}, a paradigm that was later largely superseded by velocity prediction (\vv-prediction)~\cite{salimans2022progressive,karras2022elucidating}, especially following the rise in popularity of flow matching models~\cite{lipman2022flow}.
Most recently, research into high-resolution generation---particularly for models operating directly in pixel space---has revealed that predicting the clean data (\xx-prediction) often yields superior results and enhanced stability compared to other options~\cite{hoogeboom2024simpler,li2025back,hafner2025training}.
This empirical shift suggests that as we scale models to higher resolutions and more complex data distributions, the "optimal" choice of prediction target is not static but evolves with the task.

% Despite these empirical observations, the underlying reason why different regimes favor different targets remains a fundamental open question.
% Despite these empirical advancements, the theoretical landscape remains fragmented, leaving several fundamental questions unanswered.
Despite these empirical advancements, the theoretical landscape remains fragmented, leaving several fundamental questions unanswered.
% First, it is unclear whether the optimal prediction target is strictly confined to the conventional triad of $\varepsilon$, $v$, and $x$, or if other alternatives exist for specific data regimes that have yet to be explored.
% First, it is unclear whether the optimal prediction target is strictly confined to the conventional triad of $\varepsilon$, $v$, and $x$, or if other alternatives might be superior for specific data regimes that have yet to be explored.
First, it is unclear whether the optimal prediction target is strictly confined to the conventional triad of~\ee,~\vv, and~\xx, or if superior alternatives exist within a more expansive, generalized objective space.
% For example, for low dimensional data, there is no clear answer if $v$-prediction would surpass all other candidates or not, and for high dimensional data, we are left unanswered whether $x$-prediction is a mandatory and necessary choice.
For instance, in low-dimensional settings, it remains unproven whether~\vv-prediction represents a universal optimum; conversely, for high-dimensional data, we lack a definitive answer as to whether~\xx-prediction is an indispensable requirement.
% Second, the field lacks a rigorous theoretical explanation for why data dimensionality fundamentally impacts the efficiency and stability of these targets.
Second, the field lacks a rigorous theoretical explanation for how data dimensionality fundamentally dictates the efficiency and stability of these targets.
While prior literature has invoked the manifold hypothesis to qualitatively justify~\xx-prediction for sparse, high-dimensional data~\cite{li2025back}, such conceptual frameworks are insufficient for rigorous analysis and cannot substitute for a formal quantitative description.
% This leads to a further ambiguity: whether a direct, quantifiable relationship exists between the ambient/intrinsic dimensionality of the data and the optimal prediction, and if so, whether the latter can be analytically derived from the former.
This absence of mathematical precision leads to a further ambiguity: whether a direct, quantifiable relationship exists between the data's ambient or intrinsic dimension and its optimal prediction target, and if so, whether the latter can be analytically derived from the former.
Indeed, the precise mechanism by which the optimal target shifts from~\vv-prediction toward~\xx-prediction as dimensionality increases remains unknown.
Finally, from a practical standpoint, there remains no principled methodology to determine the ideal target for a given dataset.
Currently, researchers and practitioners are forced to rely on computationally expensive heuristic-driven searches or trial-and-error, which hampers the efficiency of model development for novel high-dimensional domains.

In this work, we systematically address these gaps through a novel theoretical and algorithmic framework.
To this end, we first introduce a generic formalism that accommodates prediction targets of arbitrary functional forms, treating standard targets as specific instances of a unified objective.
To gain fundamental insights into the training process, we analyze a much simplified diffusion model, which consists of a single linear layer tasked to learn across the entire spectrum of Signal-to-Noise Ratios (SNR) of the diffusion process.
Surprisingly, such an SNR-invariant linear model exhibits remarkably rich learning dynamics, offering a tractable yet powerful lens through which to examine the interplay between data dimensionality and prediction targets.
Specifically, the optimization dynamics decouple into two distinct modes: a parallel mode aligned with the data manifold and a perpendicular mode orthogonal to it.
Notably, we demonstrate that these two modes exhibit fundamentally different learning behaviors; while the parallel mode is governed by data recovery, the perpendicular mode is dominated by noise elimination.
Since the relative contribution of these two modes is governed by the dataset's dimensionality, a fixed prediction target---chosen without regard for these geometric properties---is inherently suboptimal.

% To gain further insights into the quantitative relationship between the dimensionality and the optimal prediction target, we consider a generic form of target $\bm{u}=k\bm{x}-(1-k)\bm{\varepsilon}$, which is simple yet general enough, and could incorporate the convensional ones as special cases (i.e., $k=0, 0.5, 1$ for~\ee-,~\vv-, and~\xx-prediction, respectively).
To further investigate the quantitative relationship between dimensionality and the optimal prediction objective, we introduce a generalized target formulation: $\bm{u}=k\bm{x}-(1-k)\bm{n}$ with $0\le k\le1$.
While mathematically concise, this expression defines a continuous objective space that naturally encompasses conventional targets as discrete points along the spectrum; specifically, $k=0$, $k=0.5$, and $k=1$ recover~\ee-,~\vv-,~and~\xx-prediction, respectively (up to a constant scaling factor).
By optimizing the loss with respect to the parameter $k$, we derive the optimal target as a function of both the intrinsic and ambient dimension of the data.
This derivation establishes a formal, quantitative bridge between the geometric properties of the data manifold and the optimal prediction objective.
% By characterizing the minimized loss through the lens of these divergent dynamics, we derive a formal relationship between the data's dimensionality ratio and its unique optimal prediction target.

% \textcolor{red}{$k\bm{x}-(1-k)\bm{\varepsilon}$}

Finally, our parameterization of $\bm{u}=k\bm{x}-(1-k)\bm{n}$ also allows us to move beyond the selection of isolated candidates and instead treat the search for an optimal target as a continuous optimization problem over the scalar $k$.
% Indeed, since determining intrinsic dimension is often intractable in practice, we propose $k$-Diff, a framework that employs a data-driven method to automatically learn the optimal prediction parameter $k$ for a given dataset during training.
% Indeed, since determining the intrinsic dimension of a dataset is often intractable, we propose $k$-Diff---a framework that requires the introduction of only a single extra learnable parameter to automatically identify the optimal prediction target via standard backpropagation.
Because intrinsic dimension is difficult to estimate and the learned target can also depend on the denoiser, time sampler, and loss (Section~\ref{sec:theory-practice-bridge}), we propose $k$-Diff, which learns a scalar $k$ jointly with the denoiser through backpropagation.
Extensive experiments in both latent-space and pixel-space image generation demonstrate that $k$-Diff consistently achieves comparable or superior results to fixed-target baselines.
% , providing a principled and automated approach to optimizing generative performance across varying data scales.
By adding negligible parameter overhead, $k$-Diff provides a principled and automated approach to optimizing generative performance across varying data scales and representation spaces.

\section{Related Work}
\paragraph{Low-Dimensional Diffusion Models}
The foundations of modern diffusion models~\cite{sohl2015deep} were established through~\ee-prediction objectives~\cite{ho2020denoising,song2020score}, which focused primarily on capturing noise patterns in relatively low-resolution or low-dimensional data regimes.
Frameworks such as Denoising Diffusion Probabilistic Models (DDPM)~\cite{ho2020denoising} demonstrated the initial efficacy of this approach for image synthesis.
However, as the field transitioned toward deterministic sampling and straighter probability paths, velocity prediction (\vv-prediction) emerged as a dominant alternative within the Flow Matching~\cite{lipman2022flow} and Progressive Distillation paradigms~\cite{salimans2022progressive}.
Notably, through elaborated parameterization, the EDM framework~\cite{karras2022elucidating} leverages~\vv-prediction to ensure the neural network maintains constant output variance across the entire SNR spectrum, thereby significantly easing training convergence.
While these models improve stability and enable high-quality generation with fewer sampling steps in moderately high-dimensional spaces, the choice of target remains largely decoupled from the geometric properties of the data.

\paragraph{Pixel-Space High-Resolution Diffusion}
As generative modeling scaled to high-resolution pixel spaces, the limitations of standard noise and velocity targets became more apparent.
Recent large-scale models, particularly those operating directly in pixel space, have increasingly adopted data prediction (\xx-prediction) or heavily weighted SNR-based loss functions to maintain training stability~\cite{hoogeboom2024simpler,li2025back,hafner2025training}.
While empirical evidence suggests that~\xx-prediction is superior for handling the sparse structures and high-frequency details of high-resolution images, this shift from~\vv~to~\xx~has remained largely heuristic.
Our work provides the first quantitative bridge explaining why this transition is a natural consequence of the increasing dimensionality of the data manifold.

\paragraph{Learning and Generalization Dynamics of Diffusion Models}
A growing body of literature seeks to understand the mathematical underpinnings of diffusion models beyond their generative performance, focusing specifically on their ability to generalize versus their tendency to memorize the training set.
By investigating the inductive biases inherent in convolutional architectures,~\cite{kamb2024analytic} establishes a formal link between a model's creativity---its capacity to generate novel samples rather than merely memorizing training data---and the properties of locality and equivariance.
They provide both theoretical and empirical evidence that these architectural constraints govern how diffusion models generalize from discrete training data to the broader manifold.
Regarding the mechanics of the sampling process,~\cite{biroli2024dynamical} identifies three distinct dynamical regimes within the reverse-time generation process and characterizes two phase transitions between the regimes.
The first is the speciation transition, where the evolution moves from pure noise toward the emergence of coarse structures corresponding to a specific class, in a manner analogous to a symmetry-breaking phase transition.
This is followed by the collapse transition into a third regime where the diffusion model displays memorization of individual training samples, a phenomenon similar to condensation in a glass phase.
In parallel, to understand the acquisition of generation and memorization during training,~\cite{bonnaire2025diffusion} leverages an analysis of diffusion models using random feature networks and elucidates a transition from effective generation at early training stages to rote memorization in later phases.
This training evolution is characterized analytically as a two-timescale process, identifying a specific training window---which is proportional to the dataset size---where the model generates effectively before the onset of memorization.
Despite these foundational studies, existing literature primarily addresses the global properties of diffusion training and sampling; the specific impact of the choice of prediction target on these dynamics remains largely unexplored.

\vspace{-1em}

\paragraph{Learning and Generalization Dynamics of Neural Networks}
The study of learning and generalization in high-dimensional neural networks is deeply rooted in the analysis of their training dynamics and the geometric properties of their loss landscapes.
Seminal work on weight initialization~\cite{he2016identity,schoenholz2016deep} demonstrated that preserving signal variance across layers is essential for the stability of deep architectures.
Furthermore, the analysis of exact solutions for deep linear networks~\cite{saxe2013exact} has provided a fundamental understanding of how the spectral properties of the data determine the speed and order of feature learning.
~\cite{advani2020high} characterize the high-dimensional dynamics of generalization error by leveraging random matrix theory, revealing how the ratio of parameters to data and the signal-to-noise ratio govern the emergence of over-fitting and the trade-off between bias and variance.
Our work builds upon this tradition by extending these analytical frameworks to the generative setting, utilizing a simplified linear diffusion model to map the relationship between data dimensionality and the optimal prediction objective.

% \vspace{-2em}

\section{Theoretical Analysis of the Optimal Prediction}
The general form of the diffusion process involves the noisy input $\bm{z}$ from the diffusion process and the output or target $\bm{u}$ to be modeled by neural networks, which are defined as
% \begin{subequations}\label{eq:diffusion_process}
% \begin{align}
%     \bm{z} &= \alpha \bm{x}+\sigma\bm{n}\label{eq:diffusion_forward}\\
%     \bm{u} &= \varphi \bm{x}+\psi\bm{n}\label{eq:diffusion_backward}
% \end{align}
% \end{subequations}
\begin{equation}\label{eq:diffusion_process}
    \bm{z} = \alpha \bm{x}+\sigma\bm{n}, \quad
    \bm{u} = \varphi \bm{x}+\psi\bm{n}
\end{equation}
Here, $\bm{x}\in\mathbb{R}^D$ represents the clean data, and $\bm{n}\in\mathbb{R}^D$ is a Gaussian noise.
$\alpha$, $\sigma$, $\varphi$, $\psi$ are process parameters that characterize the diffusion process.
Typically, $\alpha$ and $\sigma$ are pre-defined, and $\varphi$ and $\psi$ are left to be designed based on the needs.
For example, for the widely-used flow matching model with $\bm{v}$-prediction, we have $\alpha=t$ and $\sigma=1-t$, where $t$ denotes the forward and backward diffusion time, and $\varphi=1$, $\psi=-1$.
On the other hand, for $\bm{x}$-prediction, we would have $\varphi=1$ and $\psi=0$.
Without loss of generality, we choose $\alpha$ as the independent variable, and the other three parameters are functions of $\alpha$.
Meanwhile, for simplicity, we omit the subscript $\alpha$ in $\bm{z}$ or $\bm{u}$ although they are functions of $\alpha$.
Note that $\bm{x}$ and $\bm{n}$ are constants that do not depend on $\alpha$.

\subsection{Problem Definition}
% To derive the optimal prediction for the diffusion models, we start our analysis by analyzing the learning dynamics of a single linear layer diffusion model on an intrinsically low dimensional data embedded in high dimension space.
To derive the optimal prediction target, we begin by investigating the learning dynamics of a single-layer linear diffusion model trained on intrinsically low-dimensional data embedded in a high-dimensional space.
% Following previous work~\cite{li2025back}, we consider the simplified case where the data $\bm{x}$ is an intrinsically low dimensional data embedded into a high dimension space by an orthogonal matrix $\bm{x}=P\widetilde{\bm{x}}$.
% Here, $\widetilde{\bm{x}}\in\mathbb{R}^d$ is the intrinsic data, and $P\in\mathbb{R}^{D\times d}$ is an arbitrary fixed sub-matrix of an orthogonal matrix of $D\times D$, so $P^\top P=I_{d\times d}$.
Following prior work~\cite{li2025back}, we consider a simplified setting where the data $\bm{x} \in \mathbb{R}^D$ is projected from an intrinsic latent $\widetilde{\bm{x}} \in \mathbb{R}^d$ via an orthogonal mapping $\bm{x} = P\widetilde{\bm{x}}$.
Here, $P \in \mathbb{R}^{D \times d}$ is an arbitrary sub-matrix of a $D \times D$ orthogonal matrix, such that $P^\top P = I_{d\times d}$.
% $d$ represents the data's intrinsic dimensionality, $D$ is the ambient dimensionality, and $I_{d\times d}$ is the identity matrix of shape $d\times d$. 
In this formulation, $d$ and $D$ denote the intrinsic and ambient dimensions of the data, respectively, while $I_{d\times d}$ represents the $d \times d$ identity matrix.
Consider a diffusion model consisting of a single linear layer $W \in \mathbb{R}^{D \times D}$.
The model estimates the target $\bm{u}$ from the noisy input $\bm{z}$ via:
\begin{equation}
    \hat{\bm{u}}=W\bm{z}
\end{equation}
% Given $\hat{\bm{u}}$, we are immediately able to solve for the estimated $\hat{\bm{x}}$ and $\hat{\bm{n}}$ by requiring
% \begin{subequations}
% \begin{align}
%     \bm{z}&=\alpha\hat{\bm{x}}+\sigma\hat{\bm{n}}\\
%     \hat{\bm{u}}&=\varphi\hat{\bm{x}}+\psi\hat{\bm{n}}
% \end{align}
% \end{subequations}
% which in turn gives the estimate for an arbitrary linear combination of $\bm{x}$ and $\bm{n}$.
% % Indeed, we have, for any such $\bm{w}$ defined as $\bm{w}=\xi\bm{x}+\eta\bm{n}$,
% To train the model, we adopt the mean square error (MSE) loss with respect to such a linear combined variable $\bm{w}$, which is related to the MSE of $\bm{u}$ by
% To train the model, instead of directly using the mean square error (MSE) loss between $\bm{u}$ and $\hat{\bm{u}}$ directly, typically we might use the MSE of another variable $\bm{w}$, which is also a linear combination of $\bm{x}$ and $\bm{n}$ (say the velocity $\bm{v}$).
% The MSE of these two types of variables ($\bm{w}$ and $\bm{u}$) are generally related by a scaling factor $\kappa$, defined as (more details in the Appendix)
% \begin{equation}\label{eq:kappa_definition}
%     \hat{\bm{w}}-\bm{w}=\kappa\cdot(\hat{\bm{u}}-\bm{u})
% \end{equation}
In practice, rather than directly minimizing the Mean Squared Error (MSE) between $\bm{u}$ and $\hat{\bm{u}}$, training often involves an alternative variable $\bm{w}$---such as the velocity $\bm{v}$---which is itself a linear combination of data $\bm{x}$ and noise $\bm{n}$.
As detailed in Appendix~\ref{app:kappa_definition}, the errors associated with these two variables are related by a scaling factor $\kappa$~\cite{kingma2023understanding}:
\begin{equation}\label{eq:kappa_definition}
    \hat{\bm{w}}-\bm{w}=\kappa\cdot(\hat{\bm{u}}-\bm{u})
\end{equation}
% Thus, when training the diffusion model with MSE loss of $\bm{w}$, the final loss is defined as
% \begin{equation}\label{eq:loss}
%     \mathcal{L}=\frac{1}{2}\int_\mathcal{I}\DIFF\alpha\E{\norm{W\bm{z}-\bm{u}}^2}
% \end{equation}
Consequently, when training the model using the MSE loss of $\bm{w}$, the objective function becomes:
\begin{equation}\label{eq:loss}
    \mathcal{L}=\frac{1}{2}\int_\mathcal{I}\DIFF\alpha\E{\norm{W\bm{z}-\bm{u}}^2}
\end{equation}
% In this expression, $\mathbb{E}$ denotes expectation over $\bm{x}$ and/or $\bm{n}$, and $\mathcal{I}$ represents the interval of $\alpha$ to take integration (which is usually the interval $[0, 1]$).
where $\mathbb{E}$ denotes expectation over $\bm{x}$ and/or $\bm{n}$, and $\mathcal{I}$ is the interval of integration (typically $[0, 1]$).
% \chaoyang{I don't understand this sentence, also, missing definition of $\DIFF$}
% Also, we have defined the integral density $\DIFF\alpha$ as
Here, we define the effective integration measure $\DIFF\alpha$ as
\begin{equation}
    \DIFF\alpha\coloneq\Diff\alpha\cdot\kappa^2
\end{equation}
% where $\Diff\alpha$ indicates the probability density of the sampling distribution for $\alpha$ (which can be $\mathcal{U}[0,1]$ or more complicated ones like logit-normal~\cite{esser2024scaling}).
where $\Diff\alpha$ is the probability measure of the sampling distribution for $\alpha$ (e.g., $\mathcal{U}[0,1]$ or logit-normal~\cite{esser2024scaling}).
% $\kappa^2$ as defined above has an effect of a loss weighting function (of $\alpha$).
The term $\kappa^2$ thus acts as a loss weighting function dependent on $\alpha$.
% facilitating the conversion between different losses (for example, from the original $\bm{u}$-loss where the loss is directly calculated with $\bm{u}$ to the $\bm{v}$-loss with velocity as usually adopted when training diffusion models.
% More details are provided in the Appendix).
% In the following, we will omit the integral interval $\mathcal{I}$ to simplify the notation unless necessary.
For brevity, we hereafter omit the integral interval $\mathcal{I}$ unless explicitly required.

% To further simplify the analysis, we would assume the intrinsic data $\widetilde{\bm{x}}$ itself is whitened, i.e.
To facilitate a tractable analysis, we assume the intrinsic data $\widetilde{\bm{x}}$ is whitened:
\begin{equation}\label{eq:white_data_assumption}
    \Sigma_{\widetilde{x}\widetilde{x}}=\E{\widetilde{\bm{x}}\widetilde{\bm{x}}^\top}=I_{d\times d}
\end{equation}
% which is a generally adopted convention in previous work when studying the learning dynamics of neural networks~\cite{saxe2013exact,advani2020high,tian2021understanding}.
This is a widely-adopted simplification in the study of neural network learning dynamics~\cite{saxe2013exact,tian2021understanding}, and we provide the analysis for general $\Sigma$ in Appendix~\ref{app:general_data_covariance}.
% Note that for the Gaussian white noise $\bm{n}$, we have similarly
Similarly, for Gaussian white noise $\bm{n}$, we have:
\begin{equation}\label{eq:white_noise_assumption}
    \Sigma_{nn}=\E{\bm{n}\bm{n}^\top}=I
\end{equation}
where $I$ denotes the $D\times D$ identity matrix.
% Now we could study the learning dynamics of the single linear layer diffusion model with this loss.
Under these assumptions, we now analyze the learning dynamics of the linear diffusion model.

\subsection{Learning Dynamics}
% As derived in Appendix~\ref{app:learning_dynamics}, with the diffusion process, single linear layer model, and the training loss defined above, the weight $W$ of the neural network would evolve during training as
As derived in Appendix~\ref{app:learning_dynamics}, given the diffusion process and the linear layer configuration defined above, the network weights $W$ evolve during training according to:
\begin{equation}\label{eq:learning_dynamics_total}
    \tau\frac{\diff W}{\diff\widetilde{t}}=-\int\DIFF\alpha\left(\alpha^2WPP^\top+\sigma^2W-\varphi\alpha PP^\top-\psi\sigma I\right)
\end{equation}
% Here, $\tau$ represents the inverse of the learning rate, and $\widetilde{t}$ denotes the training progress of the network.
where $\tau$ represents the inverse learning rate and $\widetilde{t}$ denotes the training time.
% Specifically, after noting that $PP^\top$ is a projection operator, the dynamics of $W$ can be decoupled into two modes. One is parallel to the sub-manifold of the data, and the other is perpendicular to it. In other words, we have
Notably, since $PP^\top$ is a projection operator, the dynamics of $W$ can be decoupled into two orthogonal modes: one~\textbf{parallel} to the data manifold and one~\textbf{perpendicular} to it.
\begin{subequations}\label{eq:learning_dynamics}
\begin{align}
    \tau\frac{\diff W_\parallel}{\diff\widetilde{t}}&=-\int\DIFF\alpha\Big((\alpha^2+\sigma^2)W_\parallel-(\varphi\alpha+\psi\sigma)PP^\top\Big)\label{eq:learning_dynamics_parallel}\\
    \tau\frac{\diff W_\perp}{\diff\widetilde{t}}&=-\int\DIFF\alpha\Big(\sigma^2W_\perp-\psi\sigma(I-PP^\top)\Big)\label{eq:learning_dynamics_perpendicular}
\end{align}
\end{subequations}
Specifically, we define:
% \begin{subequations}\label{eq:two_modes_of_W}
% \begin{align}
%     W_\parallel&=WPP^\top\\
%     W_\perp&=W(I-PP^\top)
% \end{align}
% \end{subequations}
\begin{equation}\label{eq:two_modes_of_W}
    W_\parallel=WPP^\top, \quad
    W_\perp=W(I-PP^\top)
\end{equation}
% represents the parallel and perpendicular modes of $W$, respectively.
which represent the parallel and perpendicular components of $W$, respectively.

Remarkably, the two modes exhibit distinct dynamics.
% Remarkably, the two modes exhibit significantly distinct dynamics.
% These two modes exhibit remarkably distinct learning dynamics.
% Indeed, although the parallel mode depends on data-related parameters like $\alpha$ and $\varphi$, these parameters are largely irrelevant to the orthogonal mode.
While the parallel mode depends on data-related parameters such as $\alpha$ and $\varphi$, these factors are largely irrelevant to the perpendicular mode.
% Physically, the parallel mode is driven by the real data and the process is like data recovery.
% On the contrary, the perpendicular mode is not data-driven, so the process mainly involves noise eliminating.
Physically, the parallel mode is driven by the underlying data distribution, resembling a process of~\textbf{data recovery}.
In contrast, the perpendicular mode is not data-driven; its dynamics are dominated by~\textbf{noise elimination} within the ambient space.
% Moreover, these two different dynamics make different requirements and implication on the parameters $\varphi$ and $\psi$, and the final design should take into account the impact of both.

Furthermore, these diverging dynamics impose different requirements on the parameters $\varphi$ and $\psi$.
An optimal design must therefore account for the competing impacts of both modes.
Based on these dynamics, we arrive at the following theorem regarding the equilibrium weights and the resulting loss.
% With the above dynamics for the two modes, as proved in the Appendix, we have the following theorem for the optimal values of the learned weights and the corresponding loss.
\begin{theorem}\label{thm:opt_loss}
    % With the loss and learning dynamics defined as in (\ref{eq:loss}) and (\ref{eq:learning_dynamics}), respectively, the optimal value of $W$ is achieved when the dynamics for both modes reach the equilibrium, which is given by
    Under the loss and learning dynamics defined in Eq.~(\ref{eq:loss}) and Eq.~(\ref{eq:learning_dynamics}), the optimal weight $W^*$ is achieved at the equilibrium of both modes, given by:
    % \begin{subequations}\label{eq:equilibirum_conditions}
    % \begin{align}
    %     W^*&=W_\parallel^*+W_\perp^*\\
    %     &=\frac{\int\DIFF\alpha(\varphi\alpha+\psi\sigma)}{\int\DIFF\alpha(\alpha^2+\sigma^2)}PP^\top+\frac{\int\DIFF\alpha\psi\sigma}{\int\DIFF\alpha\sigma^2}(I-PP^\top)\label{eq:opt_W}
    % \end{align}
    % \end{subequations}
    \begin{equation}\label{eq:equilibirum_conditions}
        W^*=W_\parallel^*+W_\perp^*        =\frac{\int\DIFF\alpha(\varphi\alpha+\psi\sigma)}{\int\DIFF\alpha(\alpha^2+\sigma^2)}PP^\top+\frac{\int\DIFF\alpha\psi\sigma}{\int\DIFF\alpha\sigma^2}(I-PP^\top)
    \end{equation}
    % Correspondingly, the optimal loss can be found by substituting the optimal weight $W$ in (\ref{eq:opt_W}) into (\ref{eq:loss}), which is %(see the Appendix for more details)
    The corresponding optimal loss is obtained by substituting $W^*$ into Eq.~(\ref{eq:loss}):
    \begin{equation}\label{eq:opt_loss}
    \scalebox{0.95}{$\Delta^*=\underbrace{\frac{1}{2}d\Big(\int\DIFF\alpha(\varphi^2+\psi^2)-\frac{\big(\int\DIFF\alpha(\varphi\alpha+\psi\sigma)\big)^2}{\int\DIFF\alpha(\alpha^2+\sigma^2)}\Big)}_{\mbox{\small \textup{Parallel Contribution}}}+\underbrace{\frac{1}{2}(D-d)\Big(\int\DIFF\alpha\psi^2-\frac{\big(\int\DIFF\alpha\psi\sigma\big)^2}{\int\DIFF\alpha\sigma^2}\Big)}_{\mbox{\small \textup{Perpendicular Contribution}}}$}
    \end{equation}
\end{theorem}
% The optimal value of $W$ is achieved when the dynamics for both modes reach the equilibrium, which is given by
% \begin{subequations}
% \begin{align}
%     W^*&=W_\parallel^*+W_\perp^*\\
%     &=\frac{\int\DIFF\alpha(\varphi\alpha+\psi\sigma)}{\int\DIFF\alpha(\alpha^2+\sigma^2)}PP^\top+\frac{\int\DIFF\alpha\psi\sigma}{\int\DIFF\alpha\sigma^2}(I-PP^\top)\label{eq:opt_W}
% \end{align}
% \end{subequations}
% Correspondingly, the optimal loss can be found by substituting the optimal weight $W$ in (\ref{eq:opt_W}) into (\ref{eq:loss}), which is (see the Appendix for more details)
% \begin{equation}\label{eq:opt_loss}
% \begin{split}
%     \Delta^*=&\frac{1}{2}d\Big(\int\DIFF\alpha(\varphi^2+\psi^2)-\frac{\big(\int\DIFF\alpha(\varphi\alpha+\psi\sigma)\big)^2}{\int\DIFF\alpha(\alpha^2+\sigma^2)}\Big)\\
%     &+\frac{1}{2}(D-d)\Big(\int\DIFF\alpha\psi^2-\frac{\big(\int\DIFF\alpha\psi\sigma\big)^2}{\int\DIFF\alpha\sigma^2}\Big)
% \end{split}
% \end{equation}

% There are several points that we would like to emphasize regarding (\ref{eq:opt_loss}).
% \paragraph{Remark.} From the optimal loss in (\ref{eq:opt_loss}), we can make several observations.
\paragraph{Remark.} Based on the optimal loss derived in Eq.~(\ref{eq:opt_loss}), we can make several key observations.
% First, the optimal loss consists of two contributions: the intra-manifold loss from the data manifold and the residual loss from the perpendicular ambient space.
% Meanwhile, the intra-manifold loss is proportional to the intrinsic dimensionality $d$ of the data, i.e., the dimensionality of the sub-manifold on which the data reside.
% On the other hand, the residual loss from the perpendicular ambient space is proportional to the co-dimensionality $(D-d)$ of the manifold.
First, the optimal loss is composed of two distinct contributions: the~\textbf{intra-manifold loss}, arising from the data manifold, and the~\textbf{residual loss}, originating from the perpendicular ambient space.
The intra-manifold loss is proportional to the~\textbf{intrinsic dimension} $d$ of the manifold on which the data reside. 
Conversely, the residual loss from the perpendicular ambient space is proportional to the~\textbf{codimension} $(D-d)$.
% Second, similar to the learning dynamics and the optimal weight, the optimal loss contributed by the intra-manifold part depends on the data-related parameters $\alpha$ and $\varphi$, yet the contribution from the perpendicular ambient space is irrelevant to these parameters.
Second, consistent with the learning dynamics and the optimal weights, the intra-manifold loss depends on the data-dependent parameters $\alpha$ and $\varphi$.
However, the contribution from the perpendicular ambient space is entirely independent of these factors.
% Third, for low dimensional case where $D\approx d$, the intra-manifold contribution dominates, while for high dimension data where $D\gg d$, the ambient part dominates.
Third, in low-dimensional regimes where $D \approx d$, the intra-manifold contribution dominates the total loss.
In contrast, for high-dimensional data where $D \gg d$, the ambient contribution becomes the primary contributor.
% Fourth, by choosing $\psi=0$, which corresponds to~\xx-prediction, the perpendicular component would vanish, leaving only the intra-manifold component for the total loss.
Fourth, by setting $\psi=0$---which corresponds to~\xx-prediction---the perpendicular contribution of the loss vanishes, leaving only the intra-manifold contribution to determine the total loss.

% Note that up till now we are dealing with the general case where the four parameters $\alpha$, $\sigma$, $\varphi$, $\psi$ can take arbitrary values.
Up to this point, our analysis has addressed the general case where the parameters $(\alpha, \sigma, \varphi, \psi)$ may take arbitrary values.
% To further understand the impact of the dimensionality $d$ and $D$, we could minimize the optimal loss in (\ref{eq:opt_loss}) with respect to the parameters $\varphi$ and $\psi$ (note that $\alpha$ and $\sigma$ are typically pre-defined), using variational methods.
To further elucidate the impact of the dimensions $d$ and $D$, one could theoretically minimize the optimal loss in Eq.~(\ref{eq:opt_loss}) with respect to $\varphi$ and $\psi$ using variational methods (assuming $\alpha$ and $\sigma$ are predefined).
% However, such a process would be too complicated, and the results would not just depend on the dimensionality $d$ and $D$, but also on many other details.
However, such a procedure is highly complex, as the results depend not only on the dimensions $d$ and $D$ but also on several auxiliary factors.
% For example, the factors that would have an impact on the loss value and the optimized parameters would include the loss we are optimizing---which would determine the parameter $\kappa$ involved in the integration, the sampling distribution of $\alpha$ we choose to learn the diffusion models, and also the integral interval $\mathcal{I}$ mentioned above over which we take the integral.
These include the choice of the optimization objective (which determines the scaling factor $\kappa$), the sampling distribution of $\alpha$ used during training, and the specific integration interval $\mathcal{I}$.
% So in the following, we will make some further simplification, aiming for more insights about how the optimal parameters or prediction targets can be determined by the dimensionality $d$ and $D$.

% Next, we will consider a special case and derive the optimal parameters (as a function of $d$ and $D$) that will minimize this optimal loss.

Therefore, in the following section, we introduce further simplifications to gain deeper insights into how the optimal prediction targets are fundamentally determined by the relationship between the dimensions $d$ and $D$.

\subsection{$k$-Parameterized Diffusion Prediction Target}
% To further understand the impact of the two dimensionalities $d$ and $D$, especially their impact on the optimal prediction, here we make further simplification, where $\varphi=k$ and $\psi=-(1-k)$ with $0\le k\le 1$ being a constant parameter that does not depend on $\alpha$. 
To further elucidate the influence of the intrinsic and ambient dimensions on the optimal prediction target, we introduce a simplification where $\varphi=k$ and $\psi=-(1-k)$.
Here, $0\le k\le 1$ is a constant parameter independent of $\alpha$.
% In other words, our network prediction will now take the form
Under this parameterization, the network prediction target takes the form:
\begin{equation}\label{eq:k_diff_u}
    \bm{u}=k\bm{x}-(1-k)\bm{n}
\end{equation}
% We will also restrict ourselves to the flow matching parameterization, where $\alpha=t$ and $\sigma=1-t$.
We further specialize our analysis to the flow matching framework, setting $\alpha=t$ and $\sigma=1-t$.
% Note that up to an extra scaling factor, the conventional~\ee-,~\vv-,~\xx-predictions are all special cases corresponding to $k=0$, $k=0.5$, and $k=1$, respectively.
Notably, up to a constant scaling factor, the conventional~\ee-,~\vv-,~\xx-predictions are recovered as special cases where $k=0$, $k=0.5$, and $k=1$, respectively.
% We have the following theorem.
This leads to the following central result:

\begin{theorem}\label{thm:opt_k}
    % Suppose we learn a diffusion process in (\ref{eq:diffusion_process}) as defined by $\alpha=t$, $\sigma=1-t$, $\varphi=k$, $\psi=-(1-k)$, with $0\le k\le 1$ a parameter that is independent of $t$ and to be determined. For the loss in (\ref{eq:loss}) and the optimal loss in (\ref{eq:opt_loss}), suppose we take the interval $\mathcal{I}$ to be $[0, 1]$, and $\alpha$ is sampled from the distribution $\mathcal{U}[0, 1]$, so $\int\DIFF\alpha=\int_0^1\diff\alpha\cdot\kappa^2$.
    Consider a diffusion process defined by $\alpha=t$ and $\sigma=1-t$.
    Let the prediction target be parameterized by $\varphi=k$ and $\psi=-(1-k)$, where $k \in [0, 1]$ is independent of $t$.
    % If we are using the original $\bm{u}$-loss, in which case $\kappa=1$, the optimal loss in (\ref{eq:opt_loss}) would be minimized by
    For the objective defined in Eq.~(\ref{eq:loss}), assuming $t \sim \mathcal{U}[0, 1]$ and a loss weighting $\kappa=1$ (i.e., $\bm{w}=\bm{u}$), the optimal loss in Eq.~(\ref{eq:opt_loss}) is minimized when:
    \begin{equation}\label{eq:opt_k_uloss}
        k^*=\frac{D}{D+d}
    \end{equation}
    % On the other hand, if we are using the $\bm{v}$-loss, which corresponds to
    % \begin{equation}\label{eq:kappa_for_vloss}
    %     \kappa=\frac{\alpha+\sigma}{\varphi\sigma-\psi\alpha}=\frac{1}{k(1-t)+(1-k)t}
    % \end{equation}
    % the optimal loss in (\ref{eq:opt_loss}) would be minimized by
    % \begin{equation}\label{eq:opt_k_vloss}
    %     k^*=\frac{\sqrt{D}}{\sqrt{D}+\sqrt{d}}
    % \end{equation}
\end{theorem}

% Theorem~\ref{thm:opt_k} gives explicit relationships between the optimized prediction target and the dimensionalities $d$ and $D$. 
Theorem~\ref{thm:opt_k} provides an explicit analytical relationship between the optimal prediction target and the dimensions $d$ and $D$.
% % Specifically, from the two expressions in (\ref{eq:opt_k_uloss}) and (\ref{eq:opt_k_vloss}),
% Specifically, from the expression in (\ref{eq:opt_k_uloss}),
% we see that when $D=d$, i.e., when the data is dense in the space, $k^*=0.5$, indicating that~\vv-prediction is the optimal for low-dimensional data.
% On the other hand, for $D\gg d$, which corresponds to data residing on a low-dimension sub-manifold, $k^*=1$, making~\xx-prediction the best option.
% For the intermediate cases, neither of these two would be the optimal, and the optimal value $k^*$ would lies in the middle between them.
Specifically, Eq.~(\ref{eq:opt_k_uloss}) reveals three distinct regimes:
\begin{itemize}[leftmargin=*, nosep]
    \item~\textbf{Low-dimensional / Dense data} ($D \approx d$): Here, $k^* \approx 0.5$, confirming that~\vv-prediction is optimal when the data fills the ambient space.
    \item~\textbf{High-dimensional / Sparse data} ($D \gg d$): In this regime, where data resides on a low-dimensional manifold, $k^* \to 1$, establishing~\xx-prediction as the superior choice.
    \item~\textbf{Intermediate regimes}: For general data distributions, neither conventional target is optimal; instead, the ideal $k^*$ lies between $0.5$ and $1$.
\end{itemize}
% Our analysis not only provide the quantitative results for the optimal $k$ given the dimensionality for the first time, but also demonstrate that there are better predictions other than~\ee-,~\vv-,~\xx-predictions for diffusion models.
Our analysis provides, to our knowledge, the first quantitative derivation of the optimal $k$ relative to data geometry. Furthermore, it demonstrates that for the vast majority of real-world datasets, the optimal prediction target lies in a continuous space between the discrete points of~\ee-,~\vv-, and~\xx-prediction.

% Insert after Theorem 3.2 and its discussion, before Section 4.
% The main document needs graphicx, needspace, and bm (for \bm).
% Put k_sweep_progressive.tex and k_sweep_progressive_four_panel.pdf in figs/.
% \Needspace{4.6in}
\subsection{Efficacy of the Closed-Form Prediction}
\label{sec:theory-practice-bridge}
Theorem~\ref{thm:opt_k} predicts $k^*=D/(D+d)$ for a time-shared linear denoiser with uniform time sampling and $\bm{u}$-MSE loss.
To test its efficacy and scope, we jointly learn $k=\mathrm{sigmoid}(w_k)$ and the denoiser on isotropic Gaussian data in $d$-dimensional subspaces of $\mathbb{R}^{D}$ ($D=128$ with eleven verified $d$ values).
For each $d$, we reuse the dataset and learn one $k$ shared across $t$.
% Figure~\ref{fig:controlled-target-study} tests the baseline, objective changes, and time conditioning.
% Companion figure for the progressive controlled study.
% Place k_sweep_progressive_four_panel.pdf in this same figs/ directory.
% \begin{figure}[!ht]
% \centering
% \includegraphics[width=\linewidth]{imgs/k_sweep_progressive_four_panel.pdf}
% \caption{Controlled sweep ($D=128$). Lines are idealized linear-model optima; markers are learned $k$.
% \textbf{Top left}: Theorem~\ref{thm:opt_k} baseline.
% \textbf{Top right}: shared linear, eight-bin linear, and MLP under uniform/$\bm{u}$-MSE.
% \textbf{Bottom left}: shared-linear sampler/loss changes.
% \textbf{Bottom right}: the same models under logit-normal/$\bm{v}$-MSE.
% The MLP has markers only; top-row vertical axes are magnified.}
% \label{fig:controlled-target-study}
% \end{figure}
\begin{figure}[!ht]
\centering
\includegraphics[width=\linewidth]{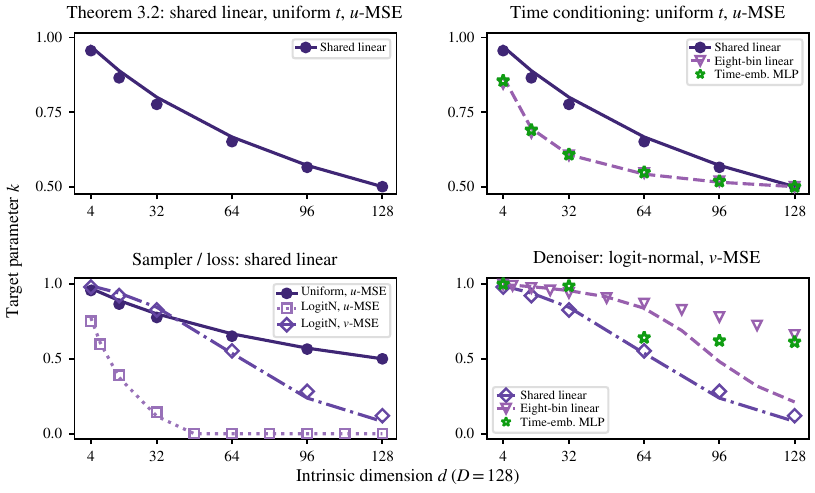}
\caption{Controlled dimension sweep ($D=128$) across model architectures, time samplers, and loss functions.
Curves represent analytical or numerical linear-model optima; markers denote empirically learned $k$ values (MLP shown with markers only).
\textbf{Top left}: Exact Theorem~\ref{thm:opt_k} baseline (shared linear model, uniform sampling, $\bm{u}$-MSE).
\textbf{Top right}: Impact of time conditioning under the baseline sampler and loss.
\textbf{Bottom left}: Effect of sampler and loss variations on the shared linear denoiser.
\textbf{Bottom right}: Combined effect of logit-normal sampling and $\bm{v}$-MSE across architectures.
Vertical axes on the top row are magnified for visual clarity.
}
\label{fig:controlled-target-study}
\end{figure}

As illustrated in Fig.~\ref{fig:controlled-target-study}, our empirical findings validate these theoretical predictions while systematically demonstrating how capacity, sampling, and loss functions shape the learned target.
Under the exact assumptions of Theorem~\ref{thm:opt_k} (top left), the learned $k$ closely tracks the closed-form prediction $D/(D+d)$ within $0.027$ across the dimension sweep.
% % Introducing time conditioning under uniform sampling and $\bm{u}$-MSE (top right) via an eight-bin linear map predicts the learned $k$ within $0.025$, while a time-embedded MLP matches the eight-bin linear model within $0.012$ (mean $0.004$).
% % This confirms that time conditioning alone accounts for the shift from the shared-linear baseline in this controlled setup.
% Introducing time conditioning while keeping uniform sampling and $\bm{u}$-MSE (top right), we assign an independent linear map to each of eight equal-width time bins.
% We first minimize the population loss over these maps, then minimize their summed loss over a single $k$ shared across bins.
% The resulting prediction matches the eight-bin model’s learned $k$ within $0.025$.
% A time-embedded MLP learns values within $0.012$ of the eight-bin model (mean difference $0.004$).
Keeping uniform sampling and $\bm{u}$-MSE, adding time conditioning (top right) shifts the learned $k$ away from $D/(D+d)$.
We use independent linear maps in eight equal-width time bins and minimize their summed population loss over one $k$ shared across bins.
This binwise prediction matches the learned eight-bin $k$ within $0.025$, while the time-embedded MLP matches the learned eight-bin model within $0.012$ (mean difference $0.004$).
Thus time conditioning explains the departure from the time-shared formula in this setting, and the binwise analysis remains predictive.
An idealized model with an independent linear map at each noise level eliminates the perpendicular residual and attains lower minimum loss than the time-shared linear model for the same fixed k (Appendix~\ref{app:time-dependent-linear}).

Modifying the objective while retaining the shared-linear denoiser (bottom left) reveals further theoretical agreement: switching to logit-normal sampling drives the optimum to $k = 0$ for $d \ge 48$ (predicted within $0.026$ by Theorem~\ref{thm:opt_loss}), whereas additionally adopting $\bm{v}$-MSE pulls the optimum away from the boundary at high $d$ (predicted within $0.045$).
Thus changing the sampler and loss moves the optimum, but the general time-shared linear analysis remains predictive.
% % Finally, combining both logit-normal sampling and $\bm{v}$-MSE (bottom right) reveals a notable divergence: at $d = 128$, the learned eight-bin model ($k = 0.659$) and MLP ($k = 0.612$) substantially exceed their idealized binwise optimum ($0.212$) and the shared-linear optimum ($0.082$).
% % % Although the binwise optimum no longer quantitatively predicts learned $k$ here, substituting the learned eight-bin $k$ into the idealized loss increases it by at most $0.121\%$ across the sweep, reflecting that unclipped $\bm{v}$-MSE becomes virtually independent of $k$ in the per-time limit (Appendix~\ref{app:controlled-target-study}).
% % This weak identification strongly motivates optimizing $k$ directly in practice.
% Finally, under logit-normal sampling and $\bm{v}$-MSE (bottom right), the eight-bin prediction no longer matches its trained model quantitatively.
% At $d = 128$, the eight-bin linear model learns $k = 0.659$, whereas its idealized population-loss optimum is $0.212$.
% The time-embedded MLP learns $k = 0.612$; for reference, the time-shared linear optimum under the same objective is $0.082$.
% The eight-bin optimum applies to that linear model and is not a theoretical optimum for the MLP.
Finally, combining time conditioning with logit-normal sampling and $\bm{v}$-MSE (bottom right), the binwise population optimum no longer quantitatively predicts its trained model, especially for large values of $d$.
For example, at $d = 128$, the eight-bin population optimum is $0.212$, whereas the trained eight-bin model learns $0.659$, and the time-embedded MLP learns $0.612$.
% : at $d = 128$, the predicted $k$ is $0.212$, whereas the eight-bin model learns $0.659$.
% The time-embedded MLP learns $0.612$; the time-shared linear optimum under the same sampler and loss is $0.082$.
% Because the discrepancy already occurs for the time-conditioned linear model, it cannot be attributed to MLP depth or nonlinearity.
% The current ablations do not isolate which part of the combined setting causes it.
% Together, these results show that knowing $d$ and $D$ alone does not prescribe the $k$ learned across denoisers and training objectives.
These results show that knowing $d$ and $D$ alone does not prescribe the $k$ learned across denoisers and training objectives.
This motivates learning $k$ under the actual training setup.

\section{Learnable Diffusion Prediction Target}
Theorem~\ref{thm:opt_k} yields $k^*=D/(D+d)$ for a time-shared linear denoiser under uniform sampling and $\bm{u}$-MSE.
However, estimating the intrinsic dimension of high-dimensional, non-linear manifolds is often intractable in practice.
Meanwhile, Section~\ref{sec:theory-practice-bridge} shows that time conditioning, sampling, and loss choice also alter the theoretical prediction.
We therefore optimize a scalar $k=\mathrm{sigmoid}(w_k)$ jointly with the denoiser instead of assigning $k$ from the closed form.
The parameters $\theta$ and $w_k$ are updated through the following training objective:
\begin{equation}
    \mathcal{L}(\theta, w_k)=\frac{1}{2}\int\DIFF\alpha\E{\norm{\bm{u}_\theta(\bm{z})-\bm{u}}^2}
    % \bm{u}=k\bm{x}-(1-k)\bm{n}
    % k=\mathrm{sigmoid}(w_k)
\end{equation}
% Section 4: replace the text after Eq. (15) through the end of Sec. 4.
% Here $\bm{u}=k\bm{x}-(1-k)\bm{n}$.
% For $\bm{v}$-MSE, the effective
% measure $\widetilde{\mathcal D}_\alpha(k)=
% \kappa(\alpha,k)^2\mathcal D_\alpha$ also depends on $k$; gradients pass through both the target and this weighting.
% We call this trainable-target method k-Diff. The learned $k$ is selected by the training objective and need not minimize a generation metric; the fixed-$k$ SWD sweep in Appendix~H
% illustrates this distinction.
% The optional time-dependent parameterization $k(t)$ and its ablation are in Appendix~F; training and sampling algorithms are in Appendix~E.
Note that here $\bm{u}$ depends on $w_k$ via $k$ as in Eq.~\ref{eq:k_diff_u}.
This end-to-end, data-driven formulation enables the model to automatically discover the optimal balance between noise elimination and data recovery.

% In addition to the simplified case of $k$ with a constant value (i.e., independent of $\alpha$) as in the derivation of the previous section, we also consider the more general case of learning an $\alpha$- or time-dependent $k(t)$.
Beyond the simplified case of a constant $k$ (time-independent), we also explore a more flexible time-dependent formulation, $k(t)$.
% For this purpose, we adopt a piecewise linear approximation method, where we divide the interval $[0, 1]$ into multiple bins (say 128 of them), and each endpoint $t_i$ of the bins is associated with one learnable parameter $w_k(t_i)$, which determine the corresponding value of $k(t_i)$ via $\mathrm{sigmoid}$.
For this purpose, we employ a piecewise linear approximation: the interval $[0, 1]$ is partitioned into $N$ bins (e.g., $N=128$), where each bin endpoint $t_i$ is associated with a learnable parameter $w_k(t_i)$.
% For any other time point $t$ inside any of the bins $[t_i, t_{i+1}]$, the value of $k(t)$ is linearly interpolated from the values $k(t_i)$ and $k(t_{i+1})$ at the two endpoints of that bin.
For any $t \in [t_i, t_{i+1}]$, the value of $k(t)$ is obtained via linear interpolation of the sigmoid-transformed endpoints (see Appendix~\ref{app:ablation_study} for more details).

% \input{tex_files/algorithm}
% To further study the 

% % We term our method as $k$-Diff, and apply it to both latent-space diffusion and pixel-space diffusion to compare with previous results using fixed output prediction. 
% We term our method $k$-Diff, and verify it on both latent-space diffusion and pixel-space diffusion.
We term our method $\bm{k}$\textbf{-Diff} and evaluate its performance in both latent-space and pixel-space diffusion regimes.
% Compared to manually tuning or defining the output prediction, our method is adaptive and has the potential to automatically determine the optimal prediction for diffusion models operating in different spaces, saving the cost for extra hyper-parameter tuning.
By adaptively determining the optimal prediction target, $k$-Diff eliminates the need for manual hyper-parameter tuning across different data domains.
% Also, our method is orthogonal to previous method for improving diffusion models, like feature alignment, complicated sampling, or advanced encoding/decoding methods, and thus can be freely plugged in to cooperate with them.
Furthermore, our approach is orthogonal to existing diffusion enhancements---such as feature alignment, advanced sampling schemes, or improved autoencoders---and can be seamlessly integrated into existing pipelines.

% We provide the pseudo-code for the training and sampling algorithms of our method in Alg.~\ref{alg:code_train} and Alg.~\ref{alg:code_sample}, respectively.
The pseudo-code for the training and sampling procedures of $k$-Diff is provided in Alg.~\ref{alg:code_train} and Alg.~\ref{alg:code_sample} in Appendix~\ref{app:algorithm}.
% For simplicity, we only provide the code for the case of a constant $k$ over the whole range of $\alpha$, i.e. the number of time bins is $1$.
% Following~\cite{li2025back}, when calculating velocity, we clamp the denominator $k * (1-t) + (1 - k) * t$ to a minimum value of $0.05$ to prevent numerical instability and division-by-zero errors.
For numerical stability, and following the practice in~\cite{li2025back}, we clamp the denominator in the velocity calculation, $k(1-t) + (1-k)t$, to a minimum value of $0.05$ to prevent division-by-zero errors when necessary (see Table~\ref{tab:config} for more details).

\section{Experiments}
In this section, we provide empirical results to verify the effectiveness of our proposed method of $k$-Diff to learn the optimal prediction.
Our experiments focus on ImageNet of resolutions 256$\times$256 and 512$\times$512.
For ImageNet~\cite{deng2009imagenet} 256$\times$256, we verify with LightningDiT-XL/1 from~\cite{yao2025reconstruction} for latent space generation and JiT-B/16 from~\cite{li2025back} for pixel space generation.
For ImageNet 512$\times$512, we only verify with JiT-B/32 from~\cite{li2025back} for pixel space generation.

% \vspace{-1em}

\subsection{Latent Space Diffusion}
% We first verify our method with latent space diffusion, mainly adopting the model and training method from~\cite{yao2025reconstruction}.
% \input{figs/vpred_vs_kDiff}
\begin{figure*}[t]
\centering
\begin{subfigure}[b]{.45\linewidth}
    \centering
    \includegraphics[width=\plotwidth]{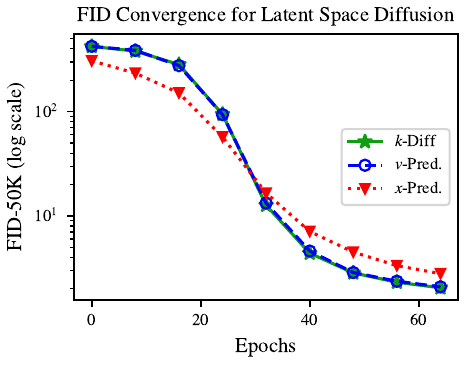}%
    \caption{Latent space diffusion.}
    \label{fig:latent_space_fid}
\end{subfigure}
% \hspace{2em}
% \hspace{1em}
\begin{subfigure}[b]{.45\linewidth}
    \centering
    \includegraphics[width=\plotwidth]{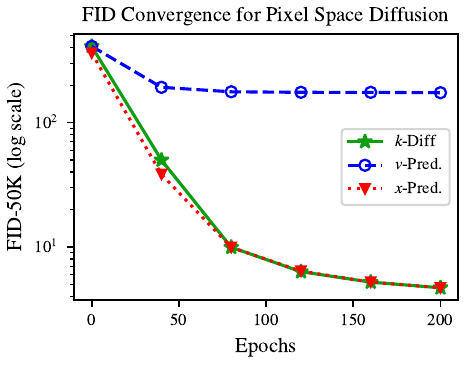}%
    \caption{Pixel space diffusion.}
    \label{fig:pixel_space_fid}
\end{subfigure}
\caption{
\textbf{FID-50k convergence dynamics of $k$-Diff in latent and pixel spaces.}
(a) Performance comparison in latent space using a LightningDiT-XL/1 architecture on ImageNet-256 over 64 epochs.
$k$-Diff maintains a convergence profile nearly identical to the original Flow Matching (\vv-prediction) framework.
(b) Convergence comparison in pixel space using a JiT-B/16 architecture over 200 epochs.
While~\xx-prediction exhibits a slightly faster initial descent, $k$-Diff recovers this gap as the learnable target stabilizes, indicating that the dynamic optimization of $k$ does not compromise training efficiency or final generative performance across disparate architectures.
}
\label{fig:fig_convergence_combined}
% \vspace{-1em}
\vspace{-.5em}
\end{figure*}
We first verify our method within the latent space diffusion regime, primarily adopting the architecture and training protocol established by~\cite{yao2025reconstruction}.
% Specifically, we train the model for 64 epochs, and sampling using CFG=1.5 with Heun solver for a number of sampling steps of 50 (NFE=99).
Specifically, we train the model for 64 epochs and perform sampling using a Classifier-Free Guidance (CFG)~\cite{ho2022classifier} scale of 1.5 with the Heun solver~\cite{heun1900neue} for 50 sampling steps (NFE=99).
% Note that this deviates a little from the setting reported in the original paper~\cite{} (CFG=10.0 with Euler Solver and 250 steps), but we find the performance with the flow matching model under the same setting as ours gives an FID of 2.08, similar to the reported one of 2.11 (for training 64 epochs).
While these hyperparameters deviate slightly from the original report \cite{yao2025reconstruction}---which utilized CFG=10.0 with an Euler solver and 250 steps---we justify this configuration by noting that the baseline Flow Matching model (\vv-prediction) yields an FID of 2.08 under our settings, which is highly consistent with the reported value of 2.11.
% On the other hand, our model achieves an FID of 2.05 after training 64 epochs.
In comparison, our model achieves a superior FID of 2.05 after the same 64 epochs of training.
% This is further illustrated in Fig.~\ref{fig:vpred_vs_kDiff_fid}, where our $k$-Diff method achieves a convergence profile similar to the original flow matching with~\vv-prediction, as measured by the FID-50k metric.
This is further illustrated in Fig.~\ref{fig:latent_space_fid}, where the $k$-Diff method exhibits a convergence profile highly similar to the original~\vv-prediction framework, as measured by the FID-50k metric across the training process.
% Specifically, our model achieves an FID of 2.05 after training 64 epochs.
% \input{figs/vpred_vs_kDiff}

\begin{figure*}[t]
\centering
\begin{subfigure}[b]{.42\linewidth}
    \centering
    \includegraphics[width=\plotwidth]{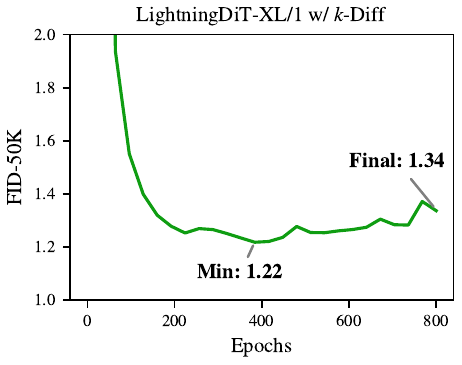}%
    \caption{Long-term FID-50k dynamics.}
    \label{fig:lightningdit_xl_1_kDiff_fid}
\end{subfigure}
% \hspace{2em}
% \hspace{1em}
\begin{subfigure}[b]{.42\linewidth}
    \centering
    \includegraphics[width=\plotwidth]{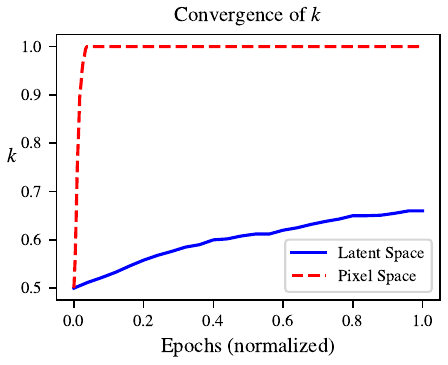}%
    \caption{Evolution of $k$.}
    \label{fig:k_convergence_normalized}
\end{subfigure}
\caption{
\textbf{Long-term FID-50k dynamics and $k$-evolution.}
(a) Latent space training (LightningDiT-XL/1) reaches a minimum FID of 1.22 at epoch 384, slightly degrading to 1.34 by epoch 800, indicating a shift in the generalization-memorization balance.
(b) Trajectories of $k$ reveal distinct optimization dynamics: in pixel space (JiT-B/16), $k$ quickly converges to 1.0 (\xx-prediction), whereas in latent space, it stabilizes at 0.66. This confirms $k$-Diff autonomously adapts to representation space, favoring intermediate targets on latent space versus~\xx-prediction in high-dimensional pixel space.
}
\label{fig:fig_convergence_combined}
% \vspace{-1em}
\vspace{-.5em}
\end{figure*}

% \input{figs/LightningDiT_XL_1_kDiff_FID}
% 1.34 for 800 epochs, both are similar to that officially reported results.
% To verify the effect of our method, we continue the training for 800 epochs, and finally achieves an FID of 1.34.
% \input{figs/LightningDiT_XL_1_kDiff_FID}
% \input{figs/latent_FID_and_k_combined}
To assess the long-term convergence of our method, we extended the training process to 800 epochs, ultimately reaching a final FID of 1.34.
As illustrated in the convergence plot of FID-50k in Fig.~\ref{fig:lightningdit_xl_1_kDiff_fid}, we observe a noticeable overfitting trend in the latter stages of training; while the final model remains highly performant, the peak generative quality is reached around epoch 384 with a minimum FID of 1.22.
Despite this intermediate peak, we maintain a consistent reporting standard: all results tabulated in Table~\ref{tab:in-sys} and the visual samples presented in Fig.~\ref{fig:imgnet_256_qualitative} are produced by the final 800-epoch checkpoint to provide a clear view of the model's stability over the entire training run.
% Meanwhile, we notice some overfitting issue for the final train model, and the best FID is achieved around epoch 384, which can reach 1.22.
% However, we only report the final results in the tables and the generated samples shown are all from the final model.
% Table~\ref{tab:latent_space} summarizes these results.
% \input{figs/k_vs_epochs_LightningDiT_XL_1}
Note that even with this slight degradation, the final FID of 1.34 remains comparable to that from the original work~\cite{yao2025reconstruction}, demonstrating the robustness of the $k$-Diff framework even when trained far beyond the point of initial convergence.

% The evolution of the FID-50k score over the full 800-epoch run reveals a rapid initial drop, followed by a prolonged period of refinement.

% It is also interesting to understand the convergence of the trained parameter $k$ as well as its final convergent value to see if the optimal results agree with flow-matching.
It is also instructive to examine the optimization trajectory of the learnable parameter $k$ and its final steady-state value to evaluate how the empirical optimum compares with the standard~\vv-prediction objective ($k=0.5$).
% As shown in Fig.~\ref{fig:k_convergence_lightningdit_xl_1}, the $k$ value gradually increases from its initial value of $0.5$, which corresponds to~\vv-prediction, to a final value of 0.66.
% As shown in Fig.~\ref{fig:k_convergence_lightningdit_xl_1},
As shown in Fig.~\ref{fig:k_convergence_normalized},
the $k$ value monotonically increases from its initial value of 0.5 to a final converged value of 0.66.
% This indicates that although the data is low-dimensional for latent space, there is still some extra dimensions and the dynamics in the extra dimensions pull the value of $k$ to a higher value.
This shift indicates that while the latent space is lower-dimensional than the pixel space, it still possesses significant ambient dimensions.
The dynamics within this orthogonal subspace pull the optimal $k$ toward a higher value, suggesting that a target closer to $x$-prediction is more effective at resolving the data manifold in this specific latent representation.

We further investigate the impact of the time-dependence of $k$ on generative quality.
As detailed in the ablation study in Appendix~\ref{app:ablation_study}, we observe that a constant $k$ is sufficient for achieving optimal performance, whereas a time-dependent parameterization using 128 bins fails to yield further improvements.
This empirical finding aligns with our theoretical analysis, suggesting that the benefits of a stable, consistent prediction objective outweigh the potential flexibility of a time-varying target.

\newcommand{\viswidth}{0.15\linewidth}
\begin{figure}[!htb]
% \begin{figure}[b]
\vspace{0.75em}
\centering
\includegraphics[width=\viswidth]{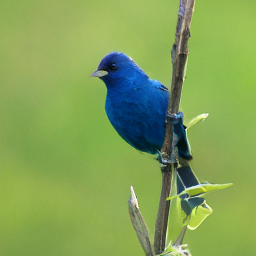}\hhs
\includegraphics[width=\viswidth]{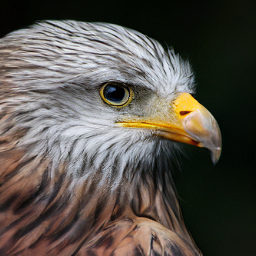}\hhs
\includegraphics[width=\viswidth]{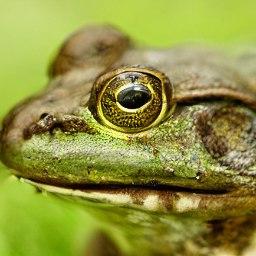}\hhs
\includegraphics[width=\viswidth]{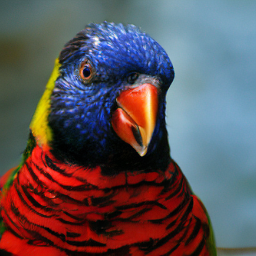}\hhs
% \includegraphics[width=\viswidth]{selected_images/029936.png}\vvs
% \\
\includegraphics[width=\viswidth]{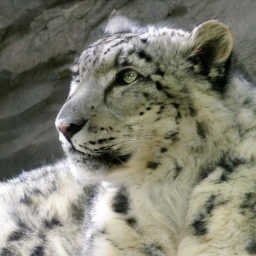}\hhs
\includegraphics[width=\viswidth]{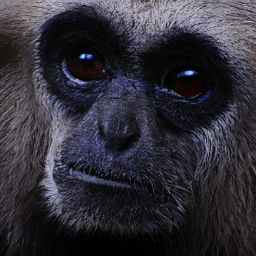}\vvs
\\
\includegraphics[width=\viswidth]{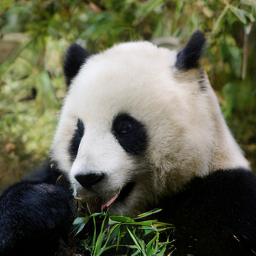}\hhs
\includegraphics[width=\viswidth]{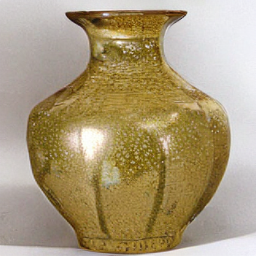}\hhs
% \includegraphics[width=\viswidth]{selected_images/024599.png}\vvs
% \\
\includegraphics[width=\viswidth]{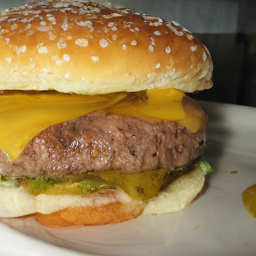}\hhs
\includegraphics[width=\viswidth]{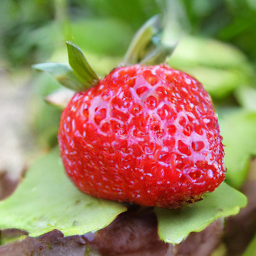}\hhs
\includegraphics[width=\viswidth]{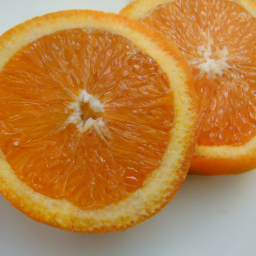}\hhs
\includegraphics[width=\viswidth]{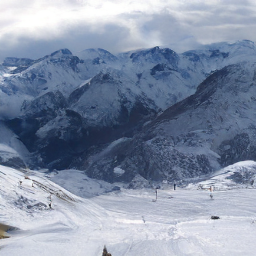}\vvs
% \includegraphics[width=\viswidth]{selected_images/046967.png}\vvs
% \\
% \includegraphics[width=\viswidth]{selected_images/022950.png}\hhs
% \includegraphics[width=\viswidth]{selected_images/018511.png}\hhs
% \includegraphics[width=\viswidth]{selected_images/042108.png}\hhs
% \includegraphics[width=\viswidth]{selected_images/004966.png}\hhs
% \includegraphics[width=\viswidth]{selected_images/046905.png}\vvs
% \\
% \includegraphics[width=\viswidth]{selected_images/034540.png}\hhs
% \includegraphics[width=\viswidth]{selected_images/027985.png}\hhs
% \includegraphics[width=\viswidth]{selected_images/007815.png}\hhs
% \includegraphics[width=\viswidth]{selected_images/019109.png}\hhs
% \includegraphics[width=\viswidth]{selected_images/012563.png}\vvs
% \\
% \includegraphics[width=\viswidth]{selected_images/047928.png}\hhs
% \includegraphics[width=\viswidth]{selected_images/040013.png}\hhs
% \includegraphics[width=\viswidth]{selected_images/012937.png}\hhs
% \includegraphics[width=\viswidth]{selected_images/011003.png}\hhs
% \includegraphics[width=\viswidth]{selected_images/035301.png}\vvs
\caption{\textbf{Qualitative Results.} Selected examples on ImageNet 256$\times$256 from LightningDiT-XL/1 trained with $k$-Diff.
}
\label{fig:imgnet_256_qualitative}
% \vspace{-2em}
% \vspace{-1em}
\vspace{-.5em}
\end{figure}

% \begin{figure}[t]
% \centering
% % \includegraphics[width=1\linewidth]{imgs/xpred_vs_kDiff_FID_plot.pdf}
% \caption{Qualitative.}
% \label{fig:imgnet_256_qualitative}
% \end{figure}
% \vspace{-30em}
In Fig.~\ref{fig:imgnet_256_qualitative}, we present a selection of generated samples produced by the model trained with the $k$-Diff objective within the latent space.
These visual results demonstrate that $k$-Diff effectively captures the complex, multi-modal distribution of the ImageNet-256 dataset.
Notably, the samples exhibit high levels of structural integrity and fine-grained textures, ranging from the intricate patterns in biological species to the sharp geometric features of architectural structures.
The quality of these images suggests that the learned target $k = 0.66$ strikes an optimal balance between denoising stability and the preservation of manifold-specific details.
% These qualitative observations complement our quantitative FID improvements, confirming that the dynamic adjustment of the prediction target enhances the generative fidelity of latent diffusion models.
% \input{figs/imgnet256_qualitative}

% \input{figs/xpred_vs_kDiff}
% \input{tabs/comparision_256}

% \input{figs/fid_convergence_combined}

% \input{figs/k_vs_epochs_normalized}

% \input{tabs/comparision_256}
% \input{tabs/comparison_512}
\newcommand\headspace{\hspace{.2em}}

\begin{table}[t]
\centering
\caption{
\textbf{Comparison results on ImageNet 256$\times$256 and 512$\times$512.} 
We evaluate our $k$-Diff objective across both latent space (using LightningDiT-XL/1) and pixel space (using JiT-B/16 and JiT-B/32) configurations.
FID \cite{heusel2017gans} and IS \cite{salimans2016improved} are computed using 50K samples.
% The ``pre-training'' columns list the external models required to obtain the results (note that the perceptual loss \cite{zhang2018unreasonable} uses a pre-trained VGG classifier \cite{simonyan2014very}).
% The parameters include the generator and tokenizer decoder (used at inference-time), but exclude other pre-trained components.
Following~\cite{li2025back}, reported parameters include the generator and tokenizer decoder required at inference time, while excluding auxiliary pre-trained modules.
% The Giga-flops are measured for a single forward pass (not counting the tokenizer) and are roughly proportional to the computational cost of an iteration during both training and inference (for the multi-scale method \cite{chen2025pixelflow}, we measure the finest level).
GFLOPs are calculated for a single forward pass—excluding the tokenizer—serving as a proxy for the computational cost per iteration during both training and inference.
For multi-scale methods \cite{chen2025pixelflow}, we measure the finest resolution level.
% \vspace{-1em}
}
\label{tab:in-sys}
\vspace{1.5pt}
\tablestyle{3pt}{1.1}
% \scriptsize
% \normalsize
% \small
\footnotesize
\setlength{\tabcolsep}{2.5pt}
% \fontsize{7.8pt}{9.4pt}\selectfont
\begin{tabular}{l | c c | c | c | l | c c | c | c}
\cellcolor{LightCyan} \textbf{ImgNet 256$\times$256} &
 \shortstack{\textbf{params}} &
 \shortstack{\textbf{Gflops}} &
 {\textbf{FID}$\downarrow$} & {\textbf{IS}$\uparrow$} &  & \shortstack{\textbf{params}} &
 \shortstack{\textbf{Gflops}} &
 {\textbf{FID}$\downarrow$} & {\textbf{IS}$\uparrow$} \\
\shline
% \rowcolor[gray]{0.9} \multicolumn{5}{l}{\textit{Latent-space Diffusion}} &  \multicolumn{5}{l}{\textit{Pixel-space Diffusion}} \\
\multicolumn{5}{l|}{\cellcolor[gray]{0.9} \textit{Latent-space Diffusion}} &  \multicolumn{5}{l}{\cellcolor[gray]{0.9} \textit{Pixel-space Diffusion}} \\
\headspace DiT-XL/2 \cite{Peebles2023} & 675+49M & 119 & 2.27 & 278.2 & \headspace ADM-G \cite{dhariwal2021diffusion} & 554M & 1120 & 4.59 & 186.7 \\
\headspace SiT-XL/2 \cite{ma2024sit} & 675+49M & 119 & 2.06 & 277.5 & \headspace RIN \cite{jabri2022scalable} & 410M & 334 & 3.42 & 182.0 \\
\headspace REPA \cite{repa}, SiT-XL/2  & 675+49M & 119 & 1.42 & 305.7 & \headspace SiD2 \cite{hoogeboom2024simpler}, UViT/2 & N/A & 137 & \textbf{1.73} & - \\
\headspace DDT-XL/2 \cite{wang2025ddt} & 675+49M & 119 & 1.26 & 310.6 & \headspace PixelFlow \cite{chen2025pixelflow}, XL/4 & 677M & 2909 & 1.98 & 282.1 \\
\headspace RAE \cite{zheng2025diffusion}, DiT$^{\text{DH}}$-XL/2 & \hspace{0pt} 839+415M & 146 & \textbf{1.13} & 262.6 & \headspace PixNerd \cite{wang2025pixnerd}, XL/16 & 700M & 134 & 2.15 & 297 \\
\cline{1-5}
\rowcolor{LightPurple} LightningDiT-XL/1 \cite{yao2025reconstruction} & 675+49M & 119 & 1.35 & 295.3 & \headspace JiT-B/16 \cite{li2025back} & 131M & 25 & 3.66 & 275.1 \\
\midline
\rowcolor{LightOrange} \headspace \textbf{k-Diff} (LightningDiT-XL/1) & 675+49M & 119 & 1.34 & 301.7 & \headspace \textbf{k-Diff} (JiT-B/16) & 131M & 25 & 3.64 & 274.9 \\
\hhline{==========}
\cellcolor{LightCyan} \textbf{ImgNet 512$\times$512} & \scriptsize\shortstack{\textbf{params}} &
 \scriptsize\shortstack{\textbf{Gflops}} &
 {\textbf{FID}$\downarrow$} & {\textbf{IS}$\uparrow$} & & \scriptsize\shortstack{\textbf{params}} &
 \scriptsize\shortstack{\textbf{Gflops}} &
 {\textbf{FID}$\downarrow$} & {\textbf{IS}$\uparrow$} \\
\shline
\multicolumn{5}{l|}{\cellcolor[gray]{0.9} \textit{Latent-space Diffusion}} & \multicolumn{5}{l}{\cellcolor[gray]{0.9} \textit{Pixel-space Diffusion}} \\
\headspace DiT-XL/2 \cite{Peebles2023} & 675+49M & 525 & 3.04 & 240.8 & \headspace RIN \cite{jabri2022scalable} & 320M & 415 & 3.95 & 216.0 \\
\headspace SiT-XL/2 \cite{ma2024sit} & 675+49M & 525 & 2.62 & 252.2 & \headspace SiD2 \cite{hoogeboom2024simpler}, UViT/4 & N/A & 137 & \textbf{2.19} & - \\
\headspace REPA \cite{repa}, SiT-XL/2  & 675+49M & 525 & 2.08 & 274.6 & \headspace PixNerd \cite{wang2025pixnerd}, XL/16 & 700 M & 583 & 2.84 & 245.6 \\
\headspace DDT-XL/2 \cite{wang2025ddt} & 675+49M & 525 & 1.28 & 305.1 & \cellcolor{LightPurple} \headspace JiT-B/32 \cite{li2025back} & \cellcolor{LightPurple} 133M & \cellcolor{LightPurple} 26 & \cellcolor{LightPurple} 4.02 & \cellcolor{LightPurple} 271.0 \\
\hhline{~~~~~-----}
% \cline{6-10}
\headspace RAE \cite{zheng2025diffusion}, DiT$^{\text{DH}}$-XL/2 & \hspace{0pt} 839+415M & 642 & \textbf{1.13} & 259.6 & \cellcolor{LightOrange} \headspace \textbf{k-Diff} (JiT-B/32) & \cellcolor{LightOrange} 133M & \cellcolor{LightOrange} 26 & \cellcolor{LightOrange} 4.03 & \cellcolor{LightOrange} 277.1
\end{tabular}

\vspace{-.5em}
\end{table}

\subsection{Pixel Space Diffusion}
% In addition to generation in latent space, we also verify our method for pixel space generation.
In addition to the latent space experiments, we evaluate the efficacy of our method within the pixel space regime.
% As shown in Fig.~\ref{fig:xpred_vs_kDiff_fid}, we compare the convergence of our method with the~\xx-prediction using JiT.
As shown in Fig.~\ref{fig:pixel_space_fid}, we compare the convergence trajectory of $k$-Diff against the~\xx-prediction baseline utilizing the JiT framework~\cite{li2025back} for ImageNet-256 with the model architecture JiT-B/16.
% We find that our method achieves comparable final convergence as the~\xx-prediction.
Our empirical results indicate that our method achieves a final convergence profile comparable to~\xx-prediction (FID=4.70 for both cases after training 200 epochs), demonstrating that $k$-Diff can adaptively recover the performance of specialized targets in high-dimensional pixel environments.

% \input{figs/imgnet256_qualitative}

% \input{figs/xpred_vs_kDiff}

% \input{tabs/comparision_256}
% \input{tabs/comparison_512}
% \input{tabs/comparison}

% This adaptive behavior is further elucidated in Fig.~\ref{fig:k_convergence_jit_b_16},
This adaptive behavior is further elucidated in Fig.~\ref{fig:k_convergence_normalized},
which tracks the evolution of the learnable parameter $k$ during the pixel-space training.
Unlike the latent space case, we observe that $k$ rapidly ascends from its initialization at 0.5 and converges to a value near 1.0 (corresponding to $x$-prediction) in fewer than 30 epochs.

This accelerated convergence suggests that the high dimensionality of the pixel space creates a strong signal for the model to prioritize data recovery over noise estimation.
The fact that $k$-Diff autonomously identifies $x$-prediction as the optimal objective confirms previous hypothesis~\cite{hoogeboom2024simpler,li2025back} and the results derived in Theorem~\ref{thm:opt_k}: in spaces where the ambient dimension $D$ is much larger than the manifold dimension $d$, the optimal target naturally shifts toward $x$.
By reaching this state so early in the training process, $k$-Diff eliminates the need for manual objective selection while ensuring the model is optimized for the correct geometric regime from the outset.

% \input{tabs/comparison}
% \input{tabs/comparision_256}
% \input{tabs/comparison_512}

% \input{figs/k_vs_epochs_JiT_B_16}

% We compare our method with previous results for both latent space diffusion and pixel space diffusion for ImageNet-256 in Tab.~\ref{tab:in256-sys} and for pixel space diffusion for ImageNet-512 in Tab.~\ref{tab:in512-sys}, respectively.
We evaluate the performance of $k$-Diff against established baselines across multiple resolutions and generation spaces.
The results for ImageNet-256, spanning both latent and pixel-space diffusion regimes, are summarized in Table~\ref{tab:in-sys}.
Furthermore, we extend our comparative analysis to pixel-space diffusion at the higher resolution of ImageNet-512.
% , as detailed in Table~\ref{tab:in512-sys}.
Across these benchmarks, our method consistently matches or outperforms existing fixed-target approaches, validating its effectiveness as a general-purpose objective for high-fidelity generative modeling.

\vspace{-1em}
\section{Conclusion and Limitation}
% In this work, we have addressed a fundamental yet often overlooked question in generative modeling: how to determine the optimal prediction target for diffusion and flow matching processes.
In this work, we addressed how to determine the optimal prediction target for diffusion and flow matching processes.
By analyzing the learning dynamics of a linear diffusion model, we provided a rigorous theoretical framework that identifies the~\textbf{intrinsic and ambient dimensions} of data as the governing factors of training efficiency.
% Our derivation of the analytical relationship, $k^* = D/(D+d)$, offers a principled justification for the empirical shift toward~\xx-prediction in high-dimensional settings while revealing that conventional targets---such as noise and velocity---are often suboptimal for the complex geometries of real-world manifolds.
Our analytical relationship, $k^* = D/(D+d)$, justifies the empirical shift toward~\xx-prediction while revealing why conventional targets are often suboptimal for real-world manifolds.
% Furthermore, our controlled sweeps demonstrate that time conditioning, samplers, and loss functions (such as unclipped $\bm{v}$-MSE, which exhibits weak identification of $k$ in the per-time limit) jointly shape the effective optimal target.
Furthermore, our controlled sweeps demonstrate that time conditioning, samplers, and loss functions jointly shape the effective optimal target.

% To bridge the gap between theory and practice, we introduced $k$-Diff, a data-driven framework that adaptively learns the optimal prediction target through standard backpropagation.
To bridge theory and practice, we introduced $k$-Diff, a framework that learns the optimal target through standard backpropagation.
% By parameterizing the target spectrum and treating it as a learnable component, $k$-Diff bypasses the intractable task of explicit dimension estimation by introducing~\textbf{only a single additional learnable scalar}.
By introducing~\textbf{only a single learnable scalar}, $k$-Diff bypasses explicit dimension estimation.
% Our experiments across both latent and pixel spaces demonstrate that $k$-Diff not only matches or exceeds the performance of manually tuned baselines but also provides a more robust and automated path for scaling generative models to novel domains.
% As diffusion models continue to grow in complexity, $k$-Diff offers a simple, model-agnostic, and theoretically grounded approach to optimizing the core objective of the generative process with negligible computational and parameter overhead.
% % By establishing that the prediction target itself can be optimized alongside the model parameters, we move closer to a fully self-configuring generative framework that eliminates the need for heuristic objective selection.
% % This work opens promising avenues for investigating how adaptive targets can be leveraged to accelerate convergence in even more challenging modalities, such as video synthesis and 3D data representations.
Experiments show $k$-Diff matches or exceeds manually tuned baselines, offering a robust, model-agnostic, and theoretically grounded path to optimize generative processes with negligible overhead.

\paragraph{Limitations.}
% % Our focus on linear models may not fully account for deep, nonlinear models or massive text-to-image tasks, which remain areas for future research.
% % However, this simplification is a deliberate choice; as Einstein noted, a theory is most impressive when it possesses "simplicity of its premises" while relating "different kinds of things".
% % By stripping away complexity, we reveal the optimal target as~\textbf{an intrinsic property of the diffusion process itself}, rather than an architectural artifact, providing a unified explanation for diverse target preferences.
% The formula $k^*=D/(D+d)$ assumes a whitened subspace distribution, a time-shared linear denoiser, uniform time sampling, and $\bm{u}$-MSE.
% Our controlled results show that changing the denoiser, sampler, or loss can change the learned target, and that a binwise population analysis does not always predict its trained value.
% Extending this analysis to nonlinear models and explaining its relationship to generation quality remain open questions.
The closed-form result $k^*=D/(D+d)$ assumes a whitened subspace distribution, a time-shared linear denoiser, uniform time sampling, and $\bm{u}$-MSE.
Our controlled experiments show that the learned target also changes with the denoiser, sampler, and loss, and that the binwise population optimum need not match the trained value.
Extending this analysis to nonlinear models and larger-scale text-to-image systems, and clarifying how target choice affects sample quality, remain open questions.

\makeatletter
\if@preprint
  \section*{Acknowledgement}
% QJ gratefully acknowledges GPUHub for their generous gift and partial support of the GPU resources utilized in this work.
QJ gratefully acknowledges GPUHub for the generous gift and the partial support of computing resources used in this study.
\else
  \if@neuripsfinal
    \section*{Acknowledgement}
% QJ gratefully acknowledges GPUHub for their generous gift and partial support of the GPU resources utilized in this work.
QJ gratefully acknowledges GPUHub for the generous gift and the partial support of computing resources used in this study.
  \fi
\fi
\makeatother

{
\small
\bibliographystyle{unsrt}
\bibliography{neurips_2026}
}

\newpage

\appendix
\onecolumn
\section{Definition of the Scaling Factor $\kappa$}\label{app:kappa_definition}
Here we provide more detailed description of the scaling factor $\kappa$ defined in Eq.~\ref{eq:kappa_definition}.
% Our analysis essentially follows the same derivation in~\cite{kingma2023understanding}.
Our analysis is aligned with the derivations presented in~\cite{kingma2023understanding}.
Generally, for the diffusion process we have $\bm{z}$ and $\bm{u}$ defined in Eq.~\ref{eq:diffusion_process}, and the estimated value from the neural network $\hat{\bm{u}}$.
Now, we could solve for the estimated $\hat{\bm{x}}$ and $\hat{\bm{n}}$ from the requirement
\begin{subequations}\label{eq:estimated_diffusion_process}
\begin{align}
    \bm{z} &= \alpha \hat{\bm{x}}+\sigma\hat{\bm{n}}\label{eq:estimated_diffusion_forward}\\
    \hat{\bm{u}} &= \varphi \hat{\bm{x}}+\psi\hat{\bm{n}}\label{eq:estimated_diffusion_backward}
\end{align}
\end{subequations}
which is similar in form to the equation of the original diffusion process given by Eq.~\ref{eq:diffusion_process}.
Indeed, from Eq.~\ref{eq:diffusion_process} and Eq.~\ref{eq:estimated_diffusion_process}, we have
\begin{subequations}
\begin{align}
    \bm{x}&=\frac{-\psi\bm{z}+\sigma\bm{u}}{\varphi\sigma-\psi\alpha}\\
    \bm{n}&=\frac{\varphi\bm{z}-\alpha\bm{u}}{\varphi\sigma-\psi\alpha}
\end{align}
\end{subequations}
and
\begin{subequations}
\begin{align}
    \hat{\bm{x}}&=\frac{-\psi\bm{z}+\sigma\hat{\bm{u}}}{\varphi\sigma-\psi\alpha}\\
    \hat{\bm{n}}&=\frac{\varphi\bm{z}-\alpha\hat{\bm{u}}}{\varphi\sigma-\psi\alpha}
\end{align}
\end{subequations}
respectively.

Now, for any target variable for MSE minimization, which is a linear combination of $\bm{x}$ and $\bm{n}$ and defined as
\begin{equation}
    \bm{w}=\xi\bm{x}+\eta\bm{n}
\end{equation}
where $\xi$ and $\eta$ are some parameters and can be functions of $\alpha$, we have its estimated value defined as
\begin{equation}
    \hat{\bm{w}}=\xi\hat{\bm{x}}+\eta\hat{\bm{n}}
\end{equation}
So the loss defined for $\bm{w}$ and that for $\bm{u}$ is related by
\begin{subequations}
\begin{align}
    \hat{\bm{w}}-\bm{w}&=\xi\left(\hat{\bm{x}}-\bm{x}\right)+\eta\left(\hat{\bm{n}}-\bm{n}\right)\\
    &=\xi\left(\frac{-\psi\bm{z}+\sigma\hat{\bm{u}}}{\varphi\sigma-\psi\alpha} - \frac{-\psi\bm{z}+\sigma\bm{u}}{\varphi\sigma-\psi\alpha}\right)+\eta\left(\frac{\varphi\bm{z}-\alpha\hat{\bm{u}}}{\varphi\sigma-\psi\alpha} - \frac{\varphi\bm{z}-\alpha\bm{u}}{\varphi\sigma-\psi\alpha}\right)\\
    &=\xi\frac{\sigma}{\varphi\sigma-\psi\alpha}\left(\hat{\bm{u}}-\bm{u}\right) + \eta\frac{-\alpha}{\varphi\sigma-\psi\alpha}\left(\hat{\bm{u}}-\bm{u}\right)\\
    &=\frac{\xi\sigma-\eta\alpha}{\varphi\sigma-\psi\alpha}\left(\hat{\bm{u}}-\bm{u}\right)\\
    &=\kappa\left(\hat{\bm{u}}-\bm{u}\right)
\end{align}
\end{subequations}
which is Eq.~\ref{eq:kappa_definition}.
Here, we have defined
\begin{equation}
    \kappa\coloneq\frac{\xi\sigma-\eta\alpha}{\varphi\sigma-\psi\alpha}
\end{equation}

Now we give some specific examples to illustrate the usage of this relationship.
For example, for the original $\bm{u}$-loss where $\bm{w}=\bm{u}$, we have
\begin{equation}
    \kappa=1
\end{equation}
For~\xx-loss, we have $\bm{w}=\bm{x}, \xi=1, \eta=0$, and
\begin{equation}
    \kappa=\frac{\sigma}{\varphi\sigma-\psi\alpha}
\end{equation}
For~\ee-loss, we have $\bm{w}=\bm{n}, \xi=0, \eta=1$, and 
\begin{equation}
    \kappa=-\frac{\alpha}{\varphi\sigma-\psi\alpha}
\end{equation}
For~\vv-loss, we have $\bm{w}=\bm{x}-\bm{n}, \xi=1, \eta=-1$, and
\begin{equation}\label{eq:kappa_for_vloss_derivation}
    \kappa=\frac{\alpha+\sigma}{\varphi\sigma-\psi\alpha}
\end{equation}
% This final expression is just Eq.~\ref{eq:kappa_for_vloss} in Theorem~\ref{thm:opt_k}.

\section{Derivation of the Learning Dynamics in Eq.~\ref{eq:learning_dynamics}}\label{app:learning_dynamics}
Here we derive the learning dynamics given in Eq.~\ref{eq:learning_dynamics}.
% We start from the loss defined in Eq.~\ref{eq:loss}.
The learning dynamics for the weight $W$ is generally given by the gradient descent
\begin{equation}
    \tau\frac{\diff W}{\diff\widetilde{t}}=-\partial_W\mathcal{L}
\end{equation}
Here, as before, $\tau$ represents the inverse of the learning rate, and $\widetilde{t}$ denotes the training progress of the network.
From the definition of the MSE loss in Eq.~\ref{eq:loss}, we have
\begin{subequations}
\begin{align}
    \tau\frac{\diff W}{\diff\widetilde{t}}&=-\partial_W\mathcal{L}\\
    &=-\partial_W\bigg[\frac{1}{2}\int\DIFF\alpha\E{\norm{W\bm{z}-\bm{u}}^2}\bigg]\\
    &=-\frac{1}{2}\int\DIFF\alpha\E{\partial_W\norm{W\bm{z}-\bm{u}}^2}\\
    &=-\frac{1}{2}\int\DIFF\alpha\E{2\left(W\bm{z}-\bm{u}\right)\bm{z}^\top}\\
    &=-\int\DIFF\alpha\bigg(W\E{\bm{z}\bm{z}^\top}-\E{\bm{u}\bm{z}^\top}\bigg)
\end{align}
\end{subequations}
where for simplicity we have omitted the integral interval $\mathcal{I}$ without ambiguity.

Now from the definition in Eq.~\ref{eq:diffusion_process}, we have
\begin{subequations}
\begin{align}
    \E{\bm{z}\bm{z}^\top}&=\E{(\alpha \bm{x}+\sigma\bm{n})(\alpha \bm{x}+\sigma\bm{n})^\top}\\
    &=\alpha^2\E{\bm{x}\bm{x}^\top}+\alpha\sigma\E{\bm{x}\bm{n}^\top}+\alpha\sigma\E{\bm{n}\bm{x}^\top}+\sigma^2\E{\bm{n}\bm{n}^\top}\\
    \E{\bm{u}\bm{z}^\top}&=\E{(\varphi \bm{x}+\psi\bm{n})(\alpha \bm{x}+\sigma\bm{n})^\top}\\
    &=\varphi\alpha\E{\bm{x}\bm{x}^\top}+\varphi\sigma\E{\bm{x}\bm{n}^\top}+\psi\alpha\E{\bm{n}\bm{x}^\top}+\psi\sigma\E{\bm{n}\bm{n}^\top}
\end{align}
\end{subequations}
Since $\bm{x}$ and $\bm{n}$ are independent random variables, and $\E{\bm{n}}=\bm{0}$, we have
\begin{subequations}
\begin{align}
    \E{\bm{x}\bm{n}^\top}&=\E{\bm{x}}\E{\bm{n}}^\top=0\\
    \E{\bm{n}\bm{x}^\top}&=\E{\bm{n}}\E{\bm{x}}^\top=0
\end{align}
\end{subequations}
From the definition $\bm{x}=P\widetilde{\bm{x}}$ and the assumptions in Eq.~\ref{eq:white_data_assumption}, we have
\begin{subequations}
\begin{align}
    \E{\bm{x}\bm{x}^\top}&=\E{P\widetilde{\bm{x}}\left(P\widetilde{\bm{x}}\right)^\top}\\
    &=P\E{\widetilde{\bm{x}}\widetilde{\bm{x}}^\top}P^\top\\
    &=PP^\top
\end{align}
\end{subequations}
Now using the white noise assumption in Eq.~\ref{eq:white_noise_assumption}, we have
\begin{subequations}\label{eq:correlation_for_zz_and_zu}
\begin{align}
    \E{\bm{z}\bm{z}^\top}&=\alpha^2PP^\top+\sigma^2 I\\
    \E{\bm{u}\bm{z}^\top}&=\varphi\alpha PP^\top+\psi\sigma I
\end{align}
\end{subequations}
The learning dynamics is thus given by
\begin{subequations}
\begin{align}
    \tau\frac{\diff W}{\diff\widetilde{t}}&=-\int\DIFF\alpha\bigg(W\E{\bm{z}\bm{z}^\top}-\E{\bm{u}\bm{z}^\top}\bigg)\\
    &=-\int\DIFF\alpha\bigg(W\left(\alpha^2PP^\top+\sigma^2 I\right)-\left(\varphi\alpha PP^\top+\psi\sigma I\right)\bigg)\\
    &=-\int\DIFF\alpha\left(\alpha^2WPP^\top+\sigma^2 W-\varphi\alpha PP^\top-\psi\sigma I\right)
\end{align}
\end{subequations}
which is exactly Eq.~\ref{eq:learning_dynamics_total}.

Now we multiply both sides of the equation on the right by $PP^\top$ and $I-PP^\top$, respectively. Notice that $PP^\top$ is a projection operation, and
\begin{subequations}
\begin{align}
    (PP^\top)(PP^\top)&=PP^\top\\
    (I-PP^\top)(I-PP^\top)&=I-PP^\top\\
    PP^\top(I-PP^\top)&=(I-PP^\top)PP^\top=0
\end{align}
\end{subequations}
we have
\begin{subequations}
\begin{align}
    \tau\frac{\diff (WPP^\top)}{\diff\widetilde{t}}&=\tau\frac{\diff W}{\diff\widetilde{t}}PP^\top\\
    &=-\int\DIFF\alpha\left(\alpha^2WPP^\top+\sigma^2 W-\varphi\alpha PP^\top-\psi\sigma I\right)PP^\top\\
    &=-\int\DIFF\alpha\left(\alpha^2WPP^\top PP^\top+\sigma^2 WPP^\top-\varphi\alpha PP^\top PP^\top-\psi\sigma PP^\top\right)\\
    &=-\int\DIFF\alpha\left(\alpha^2WPP^\top+\sigma^2 WPP^\top-\varphi\alpha PP^\top-\psi\sigma PP^\top\right)\\
    &=-\int\DIFF\alpha\left((\alpha^2+\sigma^2) WPP^\top-(\varphi\alpha+\psi\sigma) PP^\top\right)
\end{align}
\end{subequations}
and
\begin{subequations}
\begin{align}
    \tau\frac{\diff (W(I-PP^\top))}{\diff\widetilde{t}}=&\tau\frac{\diff W}{\diff\widetilde{t}}(I-PP^\top)\\
    =&-\int\DIFF\alpha\left(\alpha^2WPP^\top+\sigma^2 W-\varphi\alpha PP^\top-\psi\sigma I\right)(I-PP^\top)\\
    =&-\int\DIFF\alpha\big(\alpha^2WPP^\top(I-PP^\top)+\sigma^2 W(I-PP^\top)\nonumber\\
    &-\varphi\alpha PP^\top(I-PP^\top)-\psi\sigma (I-PP^\top)\big)\\
    =&-\int\DIFF\alpha\left(\sigma^2 W(I-PP^\top)-\psi\sigma (I-PP^\top)\right)
\end{align}
\end{subequations}
which are exactly Eq.~\ref{eq:learning_dynamics} by the definition in Eq.~\ref{eq:two_modes_of_W}.

\section{Derivation of the Optimal Loss $\Delta^*$ in Theorem~\ref{thm:opt_loss}}\label{app:opt_loss}
Now the optimal value of the two modes $W_\parallel$ and $W_\perp$ are given by the condition that the time derivatives in Eq.~\ref{eq:learning_dynamics_parallel} and Eq.~\ref{eq:learning_dynamics_perpendicular} vanish, respectively.
Notice that $W$ is not dependent on $\alpha$ and so are $W_\parallel$ and $W_\perp$, so the equilibrium conditions given by
\begin{subequations}
\begin{align}
    \int\DIFF\alpha\Big((\alpha^2+\sigma^2)W_\parallel^*-(\varphi\alpha+\psi\sigma)PP^\top\Big)&=0\\
    \int\DIFF\alpha\Big(\sigma^2W_\perp^*-\psi\sigma(I-PP^\top)\Big)&=0
\end{align}
\end{subequations}
are equivalent to
\begin{subequations}
\begin{align}
    W_\parallel^*\int\DIFF\alpha(\alpha^2+\sigma^2)-PP^\top\int\DIFF\alpha(\varphi\alpha+\psi\sigma)&=0\\
    W_\perp^*\int\DIFF\alpha\sigma^2-(I-PP^\top)\int\DIFF\alpha\psi\sigma&=0
\end{align}
\end{subequations}
which immediately gives Eq.~\ref{eq:equilibirum_conditions}.

The optimal loss is given by substituting Eq.~\ref{eq:equilibirum_conditions} into Eq.~\ref{eq:loss}, which is
\begin{subequations}
\begin{align}
    \Delta^*&=\frac{1}{2}\int\DIFF\alpha\E{\norm{W^*\bm{z}-\bm{u}}^2}\\
    &=\frac{1}{2}\int\DIFF\alpha\E{\Tr\bigg((W^*\bm{z}-\bm{u})(W^*\bm{z}-\bm{u})^\top\bigg)}\\
    &=\frac{1}{2}\int\DIFF\alpha\Tr\bigg(\E{(W^*\bm{z}-\bm{u})(W^*\bm{z}-\bm{u})^\top}\bigg)\\
    &=\frac{1}{2}\int\DIFF\alpha\Tr\bigg(\E{W^*\bm{z}\bm{z}^\top {W^*}^\top-\bm{u}\bm{z}^\top{W^*}^\top-W^*\bm{z}\bm{u}^\top+\bm{u}\bm{u}^\top}\bigg)\\
    &=\frac{1}{2}\int\DIFF\alpha\Tr\bigg(W^*\E{\bm{z}\bm{z}^\top}{W^*}^\top-\E{\bm{u}\bm{z}^\top}{W^*}^\top-W^*\E{\bm{z}\bm{u}^\top}+\E{\bm{u}\bm{u}^\top}\bigg)
\end{align}
\end{subequations}
Now as in Eq.~\ref{eq:correlation_for_zz_and_zu}, we have
\begin{subequations}
\begin{align}
    \E{\bm{z}\bm{z}^\top}&=\alpha^2PP^\top+\sigma^2 I\\
    \E{\bm{u}\bm{z}^\top}&=\varphi\alpha PP^\top+\psi\sigma I\\
    \E{\bm{z}\bm{u}^\top}&=\E{\bm{u}\bm{z}^\top}^\top\\
    &=\varphi\alpha PP^\top+\psi\sigma I\\
    \E{\bm{u}\bm{u}^\top}&=\E{(\varphi\bm{x}+\psi\bm{n})(\varphi\bm{x}+\psi\bm{n})^\top}\\
    &=\varphi^2\E{\bm{x}\bm{x}^\top}+\varphi\psi\E{\bm{x}\bm{n}^\top}+\varphi\psi\E{\bm{n}\bm{x}^\top}+\psi^2\E{\bm{n}\bm{n}^\top}\\
    &=\varphi^2PP^\top+\psi^2I
\end{align}
\end{subequations}
we have
% \begin{subequations}
\begin{align}
\label{eq:Delta_star_time_shared}
    \Delta^*=&\frac{1}{2}\int\DIFF\alpha\Tr\Big(W^*(\alpha^2PP^\top+\sigma^2 I){W^*}^\top-(\varphi\alpha PP^\top+\psi\sigma I){W^*}^\top\nonumber\\
    &-W^*(\varphi\alpha PP^\top+\psi\sigma I)+(\varphi^2PP^\top+\psi^2I)\Big)
\end{align}
% \end{subequations}
From Eq.~\ref{eq:equilibirum_conditions}, and notice that $W_\parallel^*$ and $W_\perp^*$ are directly proportional to $PP^\top$ and $I-PP^\top$, respectively, we have
\begin{subequations}
\begin{align}
    W_\parallel^*PP^\top&=PP^\top W_\parallel^*=W_\parallel^*\\
    W_\perp^*(I-PP^\top)&=(I-PP^\top)W_\perp^*=W_\perp^*\\
    W_\parallel^*(I-PP^\top)&=(I-PP^\top)W_\parallel^*=0\\
    W_\perp^*PP^\top&=PP^\top W_\perp^*=0
\end{align}
\end{subequations}
and thus
\begin{subequations}
\begin{align}
    \Tr\left(W^*PP^\top {W^*}^\top\right)&=\Tr\left((W_\parallel^*+W_\perp^*)PP^\top(W_\parallel^*+W_\perp^*)\right)\\
    &=\Tr\left(W_\parallel^*PP^\top {W_\parallel^*}^\top\right)\\
    &=\Tr\left(W_\parallel^*{W_\parallel^*}^\top\right)\\
    &=\Tr\left(\frac{\int\DIFF\alpha(\varphi\alpha+\psi\sigma)}{\int\DIFF\alpha(\alpha^2+\sigma^2)}PP^\top\frac{\int\DIFF\alpha(\varphi\alpha+\psi\sigma)}{\int\DIFF\alpha(\alpha^2+\sigma^2)}PP^\top\right)\\
    &=\frac{\left(\int\DIFF\alpha(\varphi\alpha+\psi\sigma)\right)^2}{\left(\int\DIFF\alpha(\alpha^2+\sigma^2)\right)^2}\Tr(PP^\top)
\end{align}
\end{subequations}
\begin{subequations}
\begin{align}
    \Tr\left(W^*(I-PP^\top){W^*}^\top\right)&=\Tr\left((W_\parallel^*+W_\perp^*)(I-PP^\top)(W_\parallel^*+W_\perp^*)\right)\\
    &=\Tr\left(W_\perp^*(I-PP^\top){W_\perp^*}^\top\right)\\
    &=\Tr\left(W_\perp^*{W_\perp^*}^\top\right)\\
    &=\Tr\left(\frac{\int\DIFF\alpha\psi\sigma}{\int\DIFF\alpha\sigma^2}(I-PP^\top)\frac{\int\DIFF\alpha\psi\sigma}{\int\DIFF\alpha\sigma^2}(I-PP^\top)\right)\\
    &=\frac{\left(\int\DIFF\alpha\psi\sigma\right)^2}{\left(\int\DIFF\alpha\sigma^2\right)^2}\Tr(I-PP^\top)
\end{align}
\end{subequations}
\begin{subequations}
\begin{align}
    \Tr\left(W^*PP^\top\right)&=\Tr\left(PP^\top {W^*}^\top\right)\\
    &=\Tr\left(W_\parallel^*\right)\\
    &=\frac{\int\DIFF\alpha(\varphi\alpha+\psi\sigma)}{\int\DIFF\alpha(\alpha^2+\sigma^2)}\Tr\left(PP^\top\right)
\end{align}
\end{subequations}
and
\begin{subequations}
\begin{align}
    \Tr\left(W^*(I-PP^\top)\right)&=\Tr\left((I-PP^\top){W^*}^\top\right)\\
    &=\Tr\left(W_\perp^*\right)\\
    &=\frac{\int\DIFF\alpha\psi\sigma}{\int\DIFF\alpha\sigma^2}\Tr\left(I-PP^\top\right)
\end{align}
\end{subequations}
% \begin{subequations}
% \begin{align}
%     \Tr\left(W^*\right)&=\Tr\left(W_\parallel^*+W_\perp^*\right)\\
%     &=\frac{\int\DIFF\alpha(\varphi\alpha+\psi\sigma)}{\int\DIFF\alpha(\alpha^2+\sigma^2)}\Tr\left(PP^\top\right)+\frac{\int\DIFF\alpha\psi\sigma}{\int\DIFF\alpha\sigma^2}\Tr\left(I-PP^\top\right)
% \end{align}
% \end{subequations}
With these results, we have
\begin{subequations}
\allowdisplaybreaks
\begin{align}
    \Delta^*=&\frac{1}{2}\int\DIFF\alpha\Tr\Big(W^*(\alpha^2PP^\top+\sigma^2 I){W^*}^\top-(\varphi\alpha PP^\top+\psi\sigma I){W^*}^\top\nonumber\\
    &-W^*(\varphi\alpha PP^\top+\psi\sigma I)+(\varphi^2PP^\top+\psi^2I)\Big)\\
    =&\frac{1}{2}\int\DIFF\alpha\Tr\Big(W^*(\alpha^2PP^\top+\sigma^2 I){W^*}^\top-2W^*(\varphi\alpha PP^\top+\psi\sigma I)\nonumber\\
    &+(\varphi^2PP^\top+\psi^2I)\Big)\\
    =&\frac{1}{2}\int\DIFF\alpha\Tr\Big(W^*\big((\alpha^2+\sigma^2)PP^\top+\sigma^2 (I-PP^\top)\big){W^*}^\top\nonumber\\
    &-2W^*\big((\varphi\alpha+\psi\sigma) PP^\top+\psi\sigma (I-PP^\top)\big)\nonumber\\
    &+\big((\varphi^2+\psi^2)PP^\top+\psi^2(I-PP^\top)\big)\Big)\\
    =&\frac{1}{2}\int\DIFF\alpha\Big((\alpha^2+\sigma^2)\Tr(W^*PP^\top {W^*}^\top)+\sigma^2\Tr\big(W^*(I-PP^\top){W^*}^\top\big)\nonumber\\
    &-2(\varphi\alpha+\psi\sigma)\Tr(W^*PP^\top)\nonumber\\
    &-2\psi\sigma\Tr\big(W^*(I-PP^\top)\big)+(\varphi^2+\psi^2)\Tr(PP^\top)+\psi^2\Tr(I-PP^\top)\Big)\\
    =&\frac{1}{2}\int\DIFF\alpha\Big((\alpha^2+\sigma^2)\frac{\left(\int\DIFF\bar{\alpha}(\bar{\varphi}\bar{\alpha}+\bar{\psi}\bar{\sigma})\right)^2}{\left(\int\DIFF\bar{\alpha}(\bar{\alpha}^2+\bar{\sigma}^2)\right)^2}\Tr(PP^\top)+\sigma^2\frac{\left(\int\DIFF\bar{\alpha}\bar{\psi}\bar{\sigma}\right)^2}{\left(\int\DIFF\bar{\alpha}\bar{\sigma}^2\right)^2}\Tr(I-PP^\top)\nonumber\\
    &-2(\varphi\alpha+\psi\sigma)\frac{\int\DIFF\bar{\alpha}(\bar{\varphi}\bar{\alpha}+\bar{\psi}\bar{\sigma})}{\int\DIFF\bar{\alpha}(\bar{\alpha}^2+\bar{\sigma}^2)}\Tr\left(PP^\top\right)-2\psi\sigma\frac{\int\DIFF\bar{\alpha}\bar{\psi}\bar{\sigma}}{\int\DIFF\bar{\alpha}\bar{\sigma}^2}\Tr\left(I-PP^\top\right)\nonumber\\
    &+(\varphi^2+\psi^2)\Tr(PP^\top)+\psi^2\Tr(I-PP^\top)\Big)\\
    =&\frac{1}{2}\frac{\left(\int\DIFF\alpha(\varphi\alpha+\psi\sigma)\right)^2}{\int\DIFF\alpha(\alpha^2+\sigma^2)}\Tr(PP^\top)+\frac{1}{2}\frac{\left(\int\DIFF\alpha\psi\sigma\right)^2}{\int\DIFF\alpha\sigma^2}\Tr(I-PP^\top)\nonumber\\
    &-\frac{1}{2}\cdot\frac{2\left(\int\DIFF\alpha(\varphi\alpha+\psi\sigma)\right)^2}{\int\DIFF\alpha(\alpha^2+\sigma^2)}\Tr(PP^\top)-\frac{1}{2}\cdot\frac{2\left(\int\DIFF\alpha\psi\sigma\right)^2}{\int\DIFF\alpha\sigma^2}\Tr(I-PP^\top)\nonumber\\
    &+\frac{1}{2}\int\DIFF\alpha(\varphi^2+\psi^2)\Tr(PP^\top)+\frac{1}{2}\int\DIFF\alpha\psi^2\Tr(I-PP^\top)\\
    =&\frac{1}{2}\left(\int\DIFF\alpha(\varphi^2+\psi^2)-\frac{\left(\int\DIFF\alpha(\varphi\alpha+\psi\sigma)\right)^2}{\int\DIFF\alpha(\alpha^2+\sigma^2)}\right)\Tr(PP^\top)\nonumber\\
    &+\frac{1}{2}\left(\int\DIFF\alpha\psi^2-\frac{\left(\int\DIFF\alpha\psi\sigma\right)^2}{\int\DIFF\alpha\sigma^2}\right)\Tr(I-PP^\top)
\end{align}
\end{subequations}
where we have introduced dummy variables $\bar{\alpha},\bar{\sigma},\bar{\varphi},\bar{\psi}$ to avoid symbolic ambiguity.

Notice that
\begin{subequations}
\begin{align}
    \Tr(PP^\top)&=\Tr(P^\top P)\\
    &=\Tr(I_{d\times d})\\
    &=d
\end{align}
\end{subequations}
and
\begin{subequations}
\begin{align}
    \Tr(I-PP^\top)&=\Tr(I)-\Tr(P^\top P)\\
    &=D-d
\end{align}
\end{subequations}
we have
\begin{align}
    \Delta^*=&\frac{1}{2}d\left(\int\DIFF\alpha(\varphi^2+\psi^2)-\frac{\left(\int\DIFF\alpha(\varphi\alpha+\psi\sigma)\right)^2}{\int\DIFF\alpha(\alpha^2+\sigma^2)}\right) \nonumber\\
    &+ \frac{1}{2}(D-d)\left(\int\DIFF\alpha\psi^2-\frac{\left(\int\DIFF\alpha\psi\sigma\right)^2}{\int\DIFF\alpha\sigma^2}\right)
\end{align}
which is Eq.~\ref{eq:opt_loss}.

\section{Derivation of the Optimal Value of $k$ in Theorem~\ref{thm:opt_k}}\label{app:opt_k}
Now for the simplified case where $\alpha=t,\sigma=1-t,\varphi=k,\psi=-(1-k)$ with $0\le k\le1$ independent of $t$, we have the optimal loss in Eq.~\ref{eq:opt_loss} simplified as
\begin{subequations}
\begin{align}
    \Delta^*=&\frac{1}{2}d\left(\left(k^2+(1-k)^2\right)\int\DIFF\alpha-\frac{\left(k\int\DIFF\alpha\alpha-(1-k)\int\DIFF\alpha(1-\alpha)\right)^2}{\int\DIFF\alpha\left(\alpha^2+(1-\alpha)^2\right)}\right)\nonumber\\
    &+\frac{1}{2}(D-d)\left((1-k)^2\int\DIFF\alpha-\frac{(1-k)^2\left(\int\DIFF\alpha(1-\alpha)\right)^2}{\int\DIFF\alpha(1-\alpha)^2}\right)
\end{align}
\end{subequations}
where for uniform sampling of $\alpha$ on the interval $[0, 1]$
\begin{equation}
    \int\DIFF\alpha=\int_0^1\diff\alpha\cdot\kappa^2
\end{equation}
If we use the original MSE loss of $\bm{u}$, we have $\kappa=1$, and
\begin{subequations}
\begin{align}
    \int\DIFF\alpha&=\int_0^1\diff\alpha=1\\
    \int\DIFF\alpha\alpha&=\int_0^1\diff\alpha\alpha=\frac{1}{2}\\
    \int\DIFF\alpha(1-\alpha)&=\int_0^1\diff\alpha(1-\alpha)=\frac{1}{2}\\
    \int\DIFF\alpha\alpha^2&=\int_0^1\diff\alpha\alpha^2=\frac{1}{3}\\
    \int\DIFF\alpha(1-\alpha)^2&=\int_0^1\diff\alpha(1-\alpha)^2=\frac{1}{3}
\end{align}
\end{subequations}
so we have
\begin{subequations}
\begin{align}
    \Delta^*&=\frac{1}{2}d\left(\left(k^2+(1-k)^2\right)-\frac{\left(\frac{1}{2}k-\frac{1}{2}(1-k)\right)^2}{\frac{2}{3}}\right)+\frac{1}{2}(D-d)\left((1-k)^2-\frac{(1-k)^2\cdot\left(\frac{1}{2}\right)^2}{\frac{1}{3}}\right)\\
    &=\frac{1}{2}d\left(\frac{1}{2}k^2-\frac{1}{2}k+\frac{5}{8}\right)+\frac{1}{8}(D-d)(1-k)^2\\
    &=\frac{1}{4}d\left(k^2-k+\frac{5}{4}\right)+\frac{1}{8}(D-d)(1-k)^2\\
    &=\frac{1}{16}\left[2(D+d)k^2-4Dk+(2D+3d)\right]
\end{align}
\end{subequations}
% Taking derivative, we have
% \begin{subequations}
% \begin{align}
%     \frac{\partial\Delta^*}{\partial k}&=\frac{1}{4}d(2k-1)+\frac{1}{4}(D-d)(k-1)
% \end{align}
% \end{subequations}
% which vanishes at
which is minimized by
\begin{equation}
    k^*=\frac{D}{D+d}
\end{equation}
This is the condition given in Eq.~\ref{eq:opt_k_uloss}.
% It is easy to verify that $k^*$ is a minimum since $\Delta^*$ is quadratic in $k$ and the coefficient of the quadratic term is positive.

% On the other hand, if we use the~\vv-loss, as derived in Eq.~\ref{eq:kappa_for_vloss_derivation},
% \begin{subequations}
% \begin{align}
%     \kappa&=\frac{\alpha+\sigma}{\varphi\sigma-\psi\alpha}\\
%     &=\frac{1}{k(1-t)+(1-k)t}
% \end{align}
% \end{subequations}
% we have
% \begin{subequations}
% \begin{align}
%     \int\DIFF\alpha&=\int_0^1\diff t\frac{1}{\left(k(1-t)+(1-k)t\right)^2}\\
%     &=\int_0^1\diff t\frac{1}{\left((1-2k)t+k\right)^2}\\
%     &=\frac{1}{1-2k}\int_0^{1-2k}\diff t\frac{1}{\left(t+k\right)^2}\\
%     &=\frac{1}{1-2k}\left[-\frac{1}{t+k}\right]_0^{1-2k}\\
%     &=\frac{1}{1-2k}\left(\frac{1}{k}-\frac{1}{1-2k+k}\right)\\
%     &=\frac{1}{k(1-k)}\\
%     \int\DIFF\alpha\alpha&=\int_0^1\diff t\frac{t}{\left(k(1-t)+(1-k)t\right)^2}\\
%     &=\\
%     \int\DIFF\alpha(1-\alpha)&=\int_0^1\diff t\frac{1-t}{\left(k(1-t)+(1-k)t\right)^2}\\
%     &=\\
%     \int\DIFF\alpha\alpha^2&=\int_0^1\diff t\frac{t^2}{\left(k(1-t)+(1-k)t\right)^2}\\
%     &=\\
%     \int\DIFF\alpha(1-\alpha)^2&=\int_0^1\diff t\frac{(1-t)^2}{\left(k(1-t)+(1-k)t\right)^2}\\
%     &=
% \end{align}
% \end{subequations}

\section{Training and Generation Algorithms}\label{app:algorithm}
The algorithms for training and generation with the proposed $k$-Diff method is provided in Alg.~\ref{alg:code_train} and Alg.~\ref{alg:code_sample}, respectively.
\definecolor{codeblue}{rgb}{0.25,0.5,0.5}
\definecolor{codekw}{rgb}{0.85, 0.18, 0.50}

\definecolor{codesign}{RGB}{0, 0, 255}
\definecolor{codefunc}{rgb}{0.85, 0.18, 0.50}

\lstdefinelanguage{PythonFuncColor}{
  language=Python,
  keywordstyle=\color{blue}\bfseries,
  commentstyle=\color{codeblue},
  stringstyle=\color{orange},
  showstringspaces=false,
  basicstyle=\ttfamily\small,
  literate=
    {*}{{\color{codesign}* }}{1}
    {-}{{\color{codesign}- }}{1}
    {+}{{\color{codesign}+ }}{1}
    {/}{{\color{codesign}/ }}{1}
    {dataloader}{{\color{codefunc}dataloader}}{1}
    {sample_t}{{\color{codefunc}sample\_t}}{1}
    {randn}{{\color{codefunc}randn}}{1}
    {randn_like}{{\color{codefunc}randn\_like}}{1}
    {sigmoid}{{\color{codefunc}sigmoid}}{1}
    {jvp}{{\color{codefunc}jvp}}{1}
    {stopgrad}{{\color{codefunc}stopgrad}}{1}
    {l2_loss}{{\color{codefunc}l2\_loss}}{1}
    {net_fn}{{\color{codefunc}net}}{1}
}

\lstset{
  language=PythonFuncColor,
  backgroundcolor=\color{white},
  basicstyle=\fontsize{8pt}{8.4pt}\ttfamily\selectfont,
  columns=fullflexible,
  breaklines=true,
  captionpos=b,
}

% \begin{algorithm}[b]
% \caption{Training step}
% \label{alg:code_train}
% \begin{lstlisting}
% # net(z, t): Diffusion Model
% # w_k: trainable parameter for k
% # x: training batch

% t = sample_t()
% e = randn_like(x)
% k = sigmoid(w_k)

% z = t * x + (1 - t) * e
% u = k * x - (1 - k) * e
% v = ((1 - 2 * k) * z + u) / (k * (1 - t) + (1 - k) * t)

% u_pred = net_fn(z, t)
% v_pred = ((1 - 2 * k) * z + u_pred) / (k * (1 - t) + (1 - k) * t)

% loss = l2_loss(v - v_pred)        
% \end{lstlisting}
% \end{algorithm}

% \begin{algorithm}[b]
% \caption{Sampling step (Euler)}
% \label{alg:code_sample}
% \begin{lstlisting}
% # z: current samples at t

% k = sigmoid(w_k)

% u_pred = net_fn(z, t)
% v_pred = ((1 - 2 * k) * z + u_pred) / (k * (1 - t) + (1 - k) * t)

% z_next = z + (t_next - t) * v_pred
% \end{lstlisting}
% % \vspace{-1em}
% \end{algorithm}

\begin{figure}[ht]
\centering
\begin{minipage}[t]{0.48\textwidth}
\begin{algorithm}[H]
% \begin{minipage}
\caption{Training step}
\label{alg:code_train}
\begin{lstlisting}
# net(z, t): Diffusion Model
# w_k: trainable parameter for k
# x: training batch

t = sample_t()
e = randn_like(x)
k = sigmoid(w_k)

z = t * x + (1 - t) * e
u = k * x - (1 - k) * e
v = ((1 - 2 * k) * z + u) / (k * (1 - t) + (1 - k) * t)

u_pred = net_fn(z, t)
v_pred = ((1 - 2 * k) * z + u_pred) / (k * (1 - t) + (1 - k) * t)

loss = l2_loss(v - v_pred)        
\end{lstlisting}
% \end{minipage}
\end{algorithm}
\end{minipage}
\hfill
\begin{minipage}[t]{0.48\textwidth}
\begin{algorithm}[H]
% \begin{minipage}
\caption{Sampling step (Euler)}
\label{alg:code_sample}
\begin{lstlisting}
# z: current samples at t

k = sigmoid(w_k)

u_pred = net_fn(z, t)
v_pred = ((1 - 2 * k) * z + u_pred) / (k * (1 - t) + (1 - k) * t)

z_next = z + (t_next - t) * v_pred
\end{lstlisting}
% \vspace{-1em}
% \end{minipage}
\end{algorithm}
\end{minipage}
\end{figure}

\section{Ablation Study}\label{app:ablation_study}

% \input{tabs/ablation}

% We further conduct an ablation study to justify our design choices regarding the parameterization of $k$ and the impact of the time-/$\alpha$-dependence of $k$.
In this section, we provide ablation study to justify our design choices regarding the parameterization of $k$ and the impact of the time-/$\alpha$-dependence of $k$.
% Specifically, we investigate whether increasing the degrees of freedom in the prediction target yields better generative performance.

% \input{tabs/ablation}

\paragraph{Number of independent parameters.}
We first investigate whether decoupling the coefficients for~\xx~and~\ee~improves performance by introducing more degrees of freedom.
In this setup, the target is defined as $k_0\bm{x} - k_1\bm{\varepsilon}$, where $k_0$ and $k_1$ are two independent trainable parameters both in the range of $[0, 1]$.
As shown in Table~\ref{tab:ablation}, this two-parameter approach achieves an FID of 2.04, which is nearly identical to the 2.05 achieved by our single-parameter $k$-Diff ($k\bm{x} - (1-k)\bm{\varepsilon}$).
This suggests that the relationship between the data and noise components is captured sufficiently by a single interpolation constant, and the additional complexity of a second independent parameter yields no significant generative gain.

\paragraph{Constant vs. Time-Dependent $k$.}
We also examine whether $k$ should remain constant across the diffusion process or vary with respect to the timestep $t$ (or equivalently, $\alpha$).
We implement a time-dependent version of $k$-Diff using 128 bins, where $k(t)$ is determined via linear interpolation across the range of $t$.
This configuration resulted in a degraded FID of 2.17.
We hypothesize that a constant $k$ provides a more stable optimization target, whereas a time-dependent $k$ forces the model to adapt to a shifting prediction objective at different noise levels.
This lack of target consistency may increase the difficulty of the learning task and reduce the model's robustness to estimation errors, whereas fixed-target objectives (such as~\xx,~\ee, or~\vv~in flow matching) benefit from a consistent output distribution across all timesteps.

\begin{table}[!htb]
\centering
\begin{threeparttable}
\caption{
\textbf{Ablation study on the parameterization and time-dependency of $k$.}
All models were trained for 64 epochs using the LightningDiT-XL/1 architecture on ImageNet 256$\times$256 in latent space. We compare the performance of single vs. dual independent $k$ values and the effect of using a constant $k$ versus a time-dependent $k(t)$ interpolated across 128 bins.
}
\label{tab:ablation}
\vspace{1.5pt}
\begin{tabular}{ccccc}
    \# of $k$ & Target Type & \# of Time Bins & Learned Value(s)\tnote{$\dagger$} & FID \\
    \midline
    1 ($k$-Diff) & Constant & 1 & 0.522 & 2.05 \\
    2 ($k_0, k_1$) & Constant & 1 & (0.556, 0.509) &  2.04 \\
    1 ($k$-Diff) & Time-dependent & 128 & See Fig.~\ref{fig:shift_by_one_128_k_t} & 2.17
\end{tabular}
\begin{tablenotes}
\item[$\dagger$] Learn value(s) are from Epoch 64.
\end{tablenotes}
\end{threeparttable}
\end{table}

\begin{figure}[t]
\centering
\begin{subfigure}[b]{0.45\linewidth}
\centering
\includegraphics[width=\plotwidth]{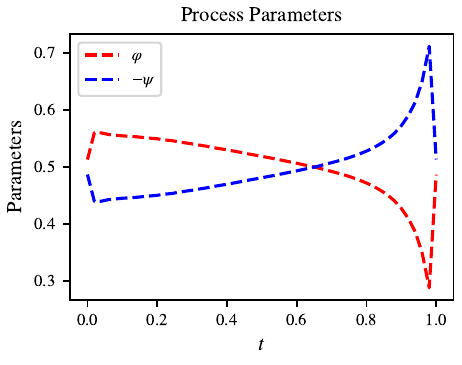}
\caption{
% \textbf{The learned time-dependent $k(t)$ for 128 time bins.}
}
\label{fig:shift_by_one_128_k_t}
\end{subfigure}
\hspace{2em}
\begin{subfigure}[b]{.45\linewidth}
\centering
\includegraphics[width=\plotwidth]{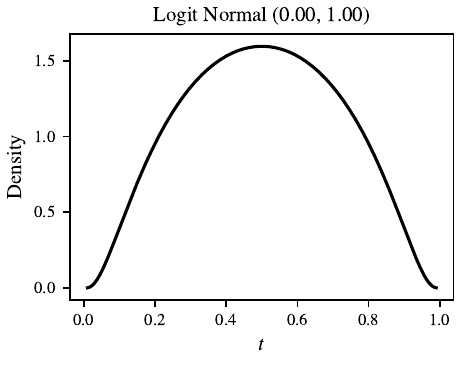}
\caption{
% \textbf{Probability distribution of the logit normal distribution with $\mu=0.0, \sigma=1.0$.}
}
\label{fig:logit_normal_0_1}
\end{subfigure}
\caption{
(a)~\textbf{The learned time-dependent $k(t)$ for 128 time bins.}
(b)~\textbf{Probability distribution of the logit normal distribution with $\mu=0.0, \sigma=1.0$.}
}
\end{figure}

% In Fig.~\ref{}, we plot the trained parameters $\varphi=k(t)$ and $-\psi=1-k(t)$ for the time-dependent case, and we find the final values are significant from the initial value (all are initialized to be 0.5).
In Fig.~\ref{fig:shift_by_one_128_k_t}, we plot the trained parameters $\varphi=k(t)$ and $-\psi=1 - k(t)$ for the time-dependent configuration, observing that the converged values deviate significantly from their initial state of 0.5.
% Meanwhile, we find that, except at the two end-points for $t=0$ and $t=1$, $k(t)$ monotonically decreases with $t$.
Excluding the boundary regions near $t=0$ and $t=1$, we find that $k(t)$ monotonically decreases as $t$ increases.
% The value of $k(t)$ start at 0.56 for $t=0$, and initially descrease slowly for small $t$, but the decreasing is much faster for large $t$, finally reaches a value near 0.3.
Specifically, $k(t)$ originates at 0.56 for $t \approx 0$ and exhibits a gradual decline in the small-$t$ regime, followed by a accelerated decreasing toward a value near 0.3 as $t$ approaches 1.
% Near the two end-points of $t=0$ and $t=1$, $k(t)$ is almost unchanged during training and remain at the initial value of $0.5$.
Crucially, $k(t)$ remains essentially stagnant at its initial value of 0.5 in the immediate vicinity of the boundaries ($t \in \{0, 1\}$).
% We guess this is due to the time-sampling distribution adopted during training.
We attribute this lack of optimization at the end-points to the density of the time-sampling distribution adopted during training.
% To this end, we plot the probability density of the logit normal distribution with $\mu=0.0$ and $\sigma=1.0$, as shown in Fig.~\ref{fig:logit_normal_0_1}.
As illustrated by the probability density function of the logit-normal distribution ($\mu=0, \sigma=1$) in Fig.~\ref{fig:logit_normal_0_1}, the sampling density vanishes at the boundaries.
% It can be seen that the probability density near the two end-points are almost zero, meaning that these two points are almost not sampled during training.
Consequently, the gradient signal at these points is insufficient to drive the parameters away from their initialization.

% \newpage
\section{Generalization to Colored Data in Ambient Space}\label{app:general_data_covariance}
% In the main text, we considered the case where data is whitened such that the covariance matrix is the identity, $I$. We now generalize these results to the case where the data follows a distribution with a general covariance matrix $\Sigma\in\mathbb{R}^{D\times D}$.
In the main text, we focused on the whitened case where the data covariance was defined by an orthogonal projection $PP^\top$. Here, we generalize the k-Diff framework to the full ambient space $\mathbb{R}^D$, where the second moment of the data is given by $\Sigma \in \mathbb{R}^{D \times D}$, i.e. $\E{\bm{x}\bm{x}^\top}=\Sigma$, which is a general symmetric positive semi-definite matrix. The case of whitened low-dimensional data analyzed in the main text is a special case where $\Sigma=PP^\top$.

In this general case, Eq.~\eqref{eq:correlation_for_zz_and_zu} is changed to
\begin{subequations}
\begin{align}
    \E{\bm{z}\bm{z}^\top}&=\alpha^2\Sigma+\sigma^2 I\\
    \E{\bm{u}\bm{z}^\top}&=\varphi\alpha\Sigma+\psi\sigma I
\end{align}
\end{subequations}
The learning dynamics is thus given by
\begin{equation}
    \tau\frac{\diff W}{\diff\widetilde{t}}=-\int\DIFF\alpha\left(\alpha^2W\Sigma+\sigma^2 W-\varphi\alpha\Sigma-\psi\sigma I\right)
\end{equation}
Now since $\Sigma$ is a symmetric semi-positive matrix, it can be decomposed into orthogonal components, each corresponding to one of the eigenvalues of $\Sigma$, i.e.
\begin{equation}
    \Sigma=\sum_i\lambda_i\bm{q}_i\bm{q}_i^\top
\end{equation}
where
\begin{equation}
    \bm{q}_j^\top\bm{q}_i=\delta_{ij}
\end{equation}
and
\begin{equation}
    \sum_i\bm{q}_i\bm{q}_i^\top=I
\end{equation}
we have the learning dynamics for each projected mode $\bm{w}_{\bm{q}_i}\coloneqq W\bm{q}_i$ given by
\begin{subequations}
\begin{align}
    \tau\frac{\diff \bm{w}_{\bm{q}_i}}{\diff\widetilde{t}}=&\tau\frac{\diff W\bm{q}_i}{\diff\widetilde{t}}\\
    =&-\int\DIFF\alpha\left(\alpha^2W\Sigma\bm{q}_i+\sigma^2 W\bm{q}_i-\varphi\alpha\Sigma\bm{q}_i-\psi\sigma I\bm{q}_i\right)\\
    =&-\int\DIFF\alpha\left(\alpha^2W\lambda_i\bm{q}_i+\sigma^2 W\bm{q}_i-\varphi\alpha\lambda_i\bm{q}_i-\psi\sigma\bm{q}_i\right)\\
    =&-\int\DIFF\alpha\left(\lambda_i\alpha^2\bm{w}_{\bm{q}_i}+\sigma^2\bm{w}_{\bm{q}_i}-\lambda_i\varphi\alpha\bm{q}_i-\psi\sigma\bm{q}_i\right)
\end{align}
\end{subequations}
which is determined by the eigenvalue $\lambda_i$. For each mode, the final optimal value at equilibrium is given by
\begin{equation}
    \bm{w}_{\bm{q}_i}^*=\frac{\int\DIFF\alpha(\lambda_i\varphi\alpha+\psi\sigma)}{\int\DIFF\alpha(\lambda_i\alpha^2+\sigma^2)}\bm{q}_i
\end{equation}
Thus the final optimal weight matrix $W^*$ is given by
\begin{subequations}
\begin{align}
    W^*=&\sum_i\bm{w}_{\bm{q}_i}^*\bm{q}_i^\top\\
    =&\sum_i\frac{\int\DIFF\alpha(\lambda_i\varphi\alpha+\psi\sigma)}{\int\DIFF\alpha(\lambda_i\alpha^2+\sigma^2)}\bm{q}_i\bm{q}_i^\top
\end{align}
\end{subequations}
The optimal loss is given by
\begin{subequations}
\begin{align}
    \Delta^*=&\frac{1}{2}\int\DIFF\alpha\E{\norm{W^*\bm{z}-\bm{u}}^2}\\
    =&\frac{1}{2}\int\DIFF\alpha\Tr\bigg(W^*\E{\bm{z}\bm{z}^\top}{W^*}^\top-\E{\bm{u}\bm{z}^\top}{W^*}^\top-W^*\E{\bm{z}\bm{u}^\top}+\E{\bm{u}\bm{u}^\top}\bigg)\\
    =&\frac{1}{2}\int\DIFF\alpha\Tr\bigg(W^*(\alpha^2\Sigma+\sigma^2I){W^*}^\top-(\varphi\alpha\Sigma+\psi\sigma I){W^*}^\top\nonumber\\
    &-W^*(\varphi\alpha\Sigma+\psi\sigma I)+(\varphi^2\Sigma+\psi^2I)\bigg)
\end{align}
\end{subequations}

Note that
\begin{subequations}
\begin{align}
    {W^*}^\top=&W^*\\
    W^*\Sigma=&\sum_i\bm{w}_{\bm{q}_i}^*\bm{q}_i^\top\Sigma\\
    =&\sum_i\bm{w}_{\bm{q}_i}^*(\Sigma\bm{q}_i)^\top\\
    =&\sum_i\bm{w}_{\bm{q}_i}^*(\lambda_i\bm{q}_i)^\top\\
    =&\sum_i\lambda_i\bm{w}_{\bm{q}_i}^*\bm{q}_i^\top\\
    =&\sum_i\lambda_i\frac{\int\DIFF\alpha(\lambda_i\varphi\alpha+\psi\sigma)}{\int\DIFF\alpha(\lambda_i\alpha^2+\sigma^2)}\bm{q}_i\bm{q}_i^\top\\
    \Sigma{W^*}^\top=&(W^*\Sigma)^\top\\
    =&W^*\Sigma\\
    W^*{W^*}^\top=&\sum_i\frac{\int\DIFF\alpha(\lambda_i\varphi\alpha+\psi\sigma)}{\int\DIFF\alpha(\lambda_i\alpha^2+\sigma^2)}\bm{q}_i\bm{q}_i^\top\sum_j\frac{\int\DIFF\alpha(\lambda_j\varphi\alpha+\psi\sigma)}{\int\DIFF\alpha(\lambda_j\alpha^2+\sigma^2)}\bm{q}_j\bm{q}_j^\top\\
    =&\sum_{ij}\frac{\int\DIFF\alpha(\lambda_i\varphi\alpha+\psi\sigma)}{\int\DIFF\alpha(\lambda_i\alpha^2+\sigma^2)}\frac{\int\DIFF\alpha(\lambda_j\varphi\alpha+\psi\sigma)}{\int\DIFF\alpha(\lambda_j\alpha^2+\sigma^2)}\bm{q}_i\bm{q}_i^\top\bm{q}_j\bm{q}_j^\top\\
    =&\sum_{ij}\frac{\int\DIFF\alpha(\lambda_i\varphi\alpha+\psi\sigma)}{\int\DIFF\alpha(\lambda_i\alpha^2+\sigma^2)}\frac{\int\DIFF\alpha(\lambda_j\varphi\alpha+\psi\sigma)}{\int\DIFF\alpha(\lambda_j\alpha^2+\sigma^2)}\bm{q}_i\delta_{ij}\bm{q}_j^\top\\
    =&\sum_{i}\left(\frac{\int\DIFF\alpha(\lambda_i\varphi\alpha+\psi\sigma)}{\int\DIFF\alpha(\lambda_i\alpha^2+\sigma^2)}\right)^2\bm{q}_i\bm{q}_i^\top\\
    W^*\Sigma{W^*}^\top=&\sum_i\lambda_i\frac{\int\DIFF\alpha(\lambda_i\varphi\alpha+\psi\sigma)}{\int\DIFF\alpha(\lambda_i\alpha^2+\sigma^2)}\bm{q}_i\bm{q}_i^\top\sum_j\frac{\int\DIFF\alpha(\lambda_j\varphi\alpha+\psi\sigma)}{\int\DIFF\alpha(\lambda_j\alpha^2+\sigma^2)}\bm{q}_j\bm{q}_j^\top\\
    =&\sum_i\lambda_i\left(\frac{\int\DIFF\alpha(\lambda_i\varphi\alpha+\psi\sigma)}{\int\DIFF\alpha(\lambda_i\alpha^2+\sigma^2)}\right)^2\bm{q}_i\bm{q}_i^\top
\end{align}
\end{subequations}
so we have
\begin{subequations}
\allowdisplaybreaks
\begin{align}
    \Delta^*=&\frac{1}{2}\int\DIFF\alpha\Tr\bigg(W^*(\alpha^2\Sigma+\sigma^2I){W^*}^\top-(\varphi\alpha\Sigma+\psi\sigma I){W^*}^\top\nonumber\\
    &-W^*(\varphi\alpha\Sigma+\psi\sigma I)+(\varphi^2\Sigma+\psi^2I)\bigg)\\
    =&\frac{1}{2}\int\DIFF\alpha\Tr\bigg(W^*(\alpha^2\Sigma+\sigma^2I){W^*}^\top-2(\varphi\alpha\Sigma+\psi\sigma I){W^*}^\top+(\varphi^2\Sigma+\psi^2I)\bigg)\\
    =&\frac{1}{2}\int\DIFF\alpha\Tr\Bigg(\alpha^2\sum_i\lambda_i\left(\frac{\int\DIFF\widetilde{\alpha}(\lambda_i\widetilde{\varphi}\widetilde{\alpha}+\widetilde{\psi}\widetilde{\sigma})}{\int\DIFF\widetilde{\alpha}(\lambda_i\widetilde{\alpha}^2+\widetilde{\sigma}^2)}\right)^2\bm{q}_i\bm{q}_i^\top+\sigma^2\sum_{i}\left(\frac{\int\DIFF\widetilde{\alpha}(\lambda_i\widetilde{\varphi}\widetilde{\alpha}+\widetilde{\psi}\widetilde{\sigma})}{\int\DIFF\widetilde{\alpha}(\lambda_i\widetilde{\alpha}^2+\widetilde{\sigma}^2)}\right)^2\bm{q}_i\bm{q}_i^\top\nonumber\\
    &-2\varphi\alpha\sum_i\lambda_i\frac{\int\DIFF\alpha(\lambda_i\varphi\alpha+\psi\sigma)}{\int\DIFF\alpha(\lambda_i\alpha^2+\sigma^2)}\bm{q}_i\bm{q}_i^\top-2\psi\sigma\sum_i\frac{\int\DIFF\alpha(\lambda_i\varphi\alpha+\psi\sigma)}{\int\DIFF\alpha(\lambda_i\alpha^2+\sigma^2)}\bm{q}_i\bm{q}_i^\top\nonumber\\
    &+(\varphi^2\Sigma+\psi^2I)\Bigg)
\end{align}
\end{subequations}
Since $\Tr{(\bm{q}_i\bm{q}_i^\top)}=\Tr{(\bm{q}_i^\top\bm{q}_i)}=1$, we have
\begin{subequations}
\begin{align}
    \Delta^*=&\frac{1}{2}\int\DIFF\alpha\Bigg(\alpha^2\sum_i\lambda_i\left(\frac{\int\DIFF\widetilde{\alpha}(\lambda_i\widetilde{\varphi}\widetilde{\alpha}+\widetilde{\psi}\widetilde{\sigma})}{\int\DIFF\widetilde{\alpha}(\lambda_i\widetilde{\alpha}^2+\widetilde{\sigma}^2)}\right)^2+\sigma^2\sum_{i}\left(\frac{\int\DIFF\widetilde{\alpha}(\lambda_i\widetilde{\varphi}\widetilde{\alpha}+\widetilde{\psi}\widetilde{\sigma})}{\int\DIFF\widetilde{\alpha}(\lambda_i\widetilde{\alpha}^2+\widetilde{\sigma}^2)}\right)^2\nonumber\\
    &-2\varphi\alpha\sum_i\lambda_i\frac{\int\DIFF\alpha(\lambda_i\varphi\alpha+\psi\sigma)}{\int\DIFF\alpha(\lambda_i\alpha^2+\sigma^2)}-2\psi\sigma\sum_i\frac{\int\DIFF\alpha(\lambda_i\varphi\alpha+\psi\sigma)}{\int\DIFF\alpha(\lambda_i\alpha^2+\sigma^2)}+(\varphi^2\Tr{\Sigma}+\psi^2\Tr{I})\Bigg)\\
    =&\frac{1}{2}\sum_i\bigg(\int\DIFF\alpha(\lambda_i\varphi^2+\psi^2)-\frac{\Big(\int\DIFF\alpha(\lambda_i\varphi\alpha+\psi\sigma)\Big)^2}{\int\DIFF\alpha(\lambda_i\alpha^2+\sigma^2)}\bigg)\\
    =&\frac{1}{2}\Tr{\bigg[\int\DIFF\alpha(\varphi^2\Sigma+\psi^2I)-\Big(\int\DIFF\alpha(\varphi\alpha\Sigma+\psi\sigma I)\Big)^2\Big(\int\DIFF\alpha(\alpha^2\Sigma+\sigma^2I)\Big)^{-1}\bigg]}
\end{align}
\end{subequations}

Note that the optimal loss can be decomposed into contribution from each mode, and for different modes, the loss and thus the corresponding optimal prediction can be different. Specifically, for the mode with eigenvalue $\lambda_i$, the contribution to the loss is given by
\begin{equation}
    \Delta_i^*=\frac{1}{2}\bigg(\int\DIFF\alpha(\lambda_i\varphi^2+\psi^2)-\frac{\Big(\int\DIFF\alpha(\lambda_i\varphi\alpha+\psi\sigma)\Big)^2}{\int\DIFF\alpha(\lambda_i\alpha^2+\sigma^2)}\bigg)
\end{equation}
For larger $\lambda_i$, i.e., larger signal-to-noise ratio (SNR), the contribution of $\varphi$ will dominate, and thus we should expect smaller $\varphi$ is better. On the other hand, for smaller $\lambda_i$, the contribution of $\psi$ dominates, and $\psi$ should be preferred. Indeed, for sufficiently large eigenvalue $\lambda_i\gg 1$, the optimal loss can be well approximated by
\begin{subequations}
\begin{align}
    \Delta_i^*\approx&\frac{1}{2}\bigg(\lambda_i\int\DIFF\alpha\varphi^2-\frac{(\lambda_i\int\DIFF\alpha\varphi\alpha)^2}{\lambda_i\int\DIFF\alpha\alpha^2}\bigg)\\
    =&\frac{1}{2}\lambda_i\bigg(\int\DIFF\alpha\varphi^2-\frac{(\int\DIFF\alpha\varphi\alpha)^2}{\int\DIFF\alpha\alpha^2}\bigg)
\end{align}
\end{subequations}
which is minimized by $\varphi=0$, i.e.,~\ee-prediction is optimal.
On the other hand, for sufficiently small eigenvalue where $\lambda_i\ll 1$, we have
\begin{subequations}
\begin{align}
    \Delta_i^*\approx&\frac{1}{2}\bigg(\int\DIFF\alpha\psi^2-\frac{(\int\DIFF\alpha\psi\sigma)^2}{\int\DIFF\alpha\sigma^2}\bigg)
\end{align}
\end{subequations}
which is minimized by $\psi=0$, i.e.,~\xx-prediction is preferred.

Under the assumptions given by Theorem~\ref{thm:opt_k}, we have
\begin{subequations}
\begin{align}
    \Delta^*=&\frac{1}{2}\sum_i\bigg(\big(k^2\lambda_i+(1-k)^2\big)-\frac{\Big(\frac{1}{2}k\lambda_i-\frac{1}{2}(1-k)\Big)^2}{\frac{1}{3}\lambda_i+\frac{1}{3}}\bigg)\\
    =&\frac{1}{2}\sum_i\bigg((1+\lambda_i)k^2-2k+1-\frac{3}{4}\frac{\big((1+\lambda_i)k-1\big)^2}{1+\lambda_i}\bigg)\\
    =&\frac{1}{2}\sum_i\bigg(\frac{1}{4}(1+\lambda_i)k^2-2k+1+\frac{3}{2}k-\frac{3}{4}\frac{1}{1+\lambda_i}\bigg)\\
    =&\frac{1}{8}\sum_i\bigg((1+\lambda_i)k^2-2k+\frac{1+4\lambda_i}{1+\lambda_i}\bigg)\\
    =&\frac{1}{8}\bigg((D+\Tr{\Sigma})k^2-2Dk+\Tr{\big((I+4\Sigma)(I+\Sigma)^{-1}\big)}\bigg)
\end{align}
\end{subequations}
The optimal prediction target is thus given by
\begin{equation}
    k^*=\frac{D}{D+\Tr\Sigma}
\end{equation}
For the special case where $\Sigma=PP^T$ as that in the main text, we have $\Tr\Sigma=d$, which agrees with the result derived in Theorem~\ref{thm:opt_k}.

% Insert among the technical appendices, preferably after the existing
% derivations and before implementation details. The equation numbers (4),
% (12), and (14) refer to the current main-text numbering; replace these
% numerical references with your existing labels if available.
\section{Noise-Level-Dependent Linear Analysis}
\label{app:time-dependent-linear}

To understand the theoretical benefits of model capacity across noise levels, we extend the analysis of Theorem~\ref{thm:opt_loss} by allowing the linear denoiser to dynamically adapt to the noise level, i.e., $\widehat{\bm{u}}=W(\alpha)\bm{z}$.
Under this noise-level-dependent setting, the learning dynamics of the weight matrix $W(\alpha)$ are governed by the functional derivative with respect to each $\alpha$:
\begin{equation}
    \tau\frac{\diff W(\alpha)}{\diff\widetilde{t}}=-\frac{\delta\mathcal{L}}{\delta W(\alpha)}
\end{equation}
By the definition of the MSE loss in Eq.~\ref{eq:loss}, we have
\begin{subequations}
\begin{align}
    \tau\frac{\diff W(\alpha)}{\diff\widetilde{t}}&=-\frac{\delta\mathcal{L}}{\delta W(\alpha)}\\
    &=-\frac{\delta}{\delta W(\alpha)}\bigg[\frac{1}{2}\int\DIFF\alpha\E{\norm{W(\alpha)\bm{z}-\bm{u}}^2}\bigg]\\
    &=-\frac{1}{2}\frac{\DIFF\alpha}{\diff\alpha}\cdot\E{\partial_{W(\alpha)}\norm{W(\alpha)\bm{z}-\bm{u}}^2}\\
    &=-\frac{1}{2}\frac{\DIFF\alpha}{\diff\alpha}\cdot\E{2\left(W(\alpha)\bm{z}-\bm{u}\right)\bm{z}^\top}\\
    &=-\frac{\DIFF\alpha}{\diff\alpha}\cdot\bigg(W(\alpha)\E{\bm{z}\bm{z}^\top}-\E{\bm{u}\bm{z}^\top}\bigg)
\end{align}
\end{subequations}
where $\frac{\tilde{\mathcal{D}}\alpha}{d\alpha}$ acts as the local density measure over the diffusion trajectory.
Substituting the statistical expectations from Eq.~\ref{eq:correlation_for_zz_and_zu}, the learning dynamics is simplified as
\begin{subequations}
\begin{align}
    \tau\frac{\diff W(\alpha)}{\diff\widetilde{t}}&=-\frac{\DIFF\alpha}{\diff\alpha}\cdot\bigg(W(\alpha)\E{\bm{z}\bm{z}^\top}-\E{\bm{u}\bm{z}^\top}\bigg)\\
    &=-\frac{\DIFF\alpha}{\diff\alpha}\cdot\bigg(W(\alpha)\left(\alpha^2PP^\top+\sigma^2 I\right)-\left(\varphi\alpha PP^\top+\psi\sigma I\right)\bigg)\\
    &=-\frac{\DIFF\alpha}{\diff\alpha}\cdot\left(\alpha^2W(\alpha)PP^\top+\sigma^2 W(\alpha)-\varphi\alpha PP^\top-\psi\sigma I\right)
\end{align}
\end{subequations}

Following the same derivation as for the time-shared linear model, the optimal weights under the equilibrium condition is given by
\begin{subequations}
\begin{align}
    W^*(\alpha)&=W_\parallel^*(\alpha)+W_\perp^*(\alpha)\\
    &=\frac{\varphi\alpha+\psi\sigma}{\alpha^2+\sigma^2}PP^T+\frac{\psi}{\sigma}(I-PP^T)
\end{align}
\end{subequations}
As in Eq.~\ref{eq:Delta_star_time_shared}, the optimal loss is
\begin{subequations}
\begin{align}
\label{eq:Delta_star_time_dependent}
    \Delta^*_\texttt{dep}=&\frac{1}{2}\int\DIFF\alpha\Tr\Big(W^*(\alpha)(\alpha^2PP^\top+\sigma^2 I){W^*(\alpha)}^\top-(\varphi\alpha PP^\top+\psi\sigma I){W^*(\alpha)}^\top\nonumber\\
    &-W^*(\alpha)(\varphi\alpha PP^\top+\psi\sigma I)+(\varphi^2PP^\top+\psi^2I)\Big)\\
    =&\frac{1}{2}\int\DIFF\alpha\Big((\varphi^2+\psi^2)-\frac{(\varphi\alpha+\psi\sigma)^2}{\alpha^2+\sigma^2}\Big)\Tr{(PP^T)}\\
    =&\frac{1}{2}d\int\DIFF\alpha\frac{(\varphi\sigma-\psi\alpha)^2}{\alpha^2+\sigma^2}
\end{align}
\end{subequations}

\paragraph{Comparison with the Time-Shared Model}
Comparing $\Delta_{\text{dep}}^{*}$ with the optimal loss of the time-shared model ($\Delta^{*}$ in Eq.~\ref{eq:opt_loss}), we have several observations.
First, unlike the time-shared model, whose optimal loss contains a persistent ambient penalty proportional to $(D-d)$, the noise-level-dependent loss $\Delta^*_{\texttt{dep}}$ collapses entirely into the intra-manifold term.
This demonstrates that granting the linear model noise-level dependence completely eliminates the perpendicular residual penalty.
Additionally, we can establish a direct analytical relationship using integral inequalities.
Specifically, by applying Cauchy-Schwarz-type inequalities:
\begin{subequations}
\begin{align}
    \int\DIFF\alpha\frac{(\varphi\alpha+\psi\sigma)^2}{\alpha^2+\sigma^2}\int\DIFF\alpha(\alpha^2+\sigma^2)&\ge\big(\int\DIFF\alpha(\varphi\alpha+\psi\sigma)\big)^2\\
    \int\DIFF\alpha\psi^2\int\DIFF\alpha\sigma^2&\ge\big(\int\DIFF\alpha\psi\sigma\big)^2
\end{align}
\end{subequations}
we can see that
\begin{subequations}
\begin{align}
    \Delta^*_\texttt{dep}=&\frac{1}{2}\int\DIFF\alpha\Big((\varphi^2+\psi^2)-\frac{(\varphi\alpha+\psi\sigma)^2}{\alpha^2+\sigma^2}\Big)\Tr{(PP^T)}\\
    \le&\frac{1}{2}\left(\int\DIFF\alpha(\varphi^2+\psi^2)-\frac{\left(\int\DIFF\alpha(\varphi\alpha+\psi\sigma)\right)^2}{\int\DIFF\alpha(\alpha^2+\sigma^2)}\right)\Tr(PP^\top)\\
    \le&\frac{1}{2}\left(\int\DIFF\alpha(\varphi^2+\psi^2)-\frac{\left(\int\DIFF\alpha(\varphi\alpha+\psi\sigma)\right)^2}{\int\DIFF\alpha(\alpha^2+\sigma^2)}\right)\Tr(PP^\top)\nonumber\\
    &+\frac{1}{2}\left(\int\DIFF\alpha\psi^2-\frac{\left(\int\DIFF\alpha\psi\sigma\right)^2}{\int\DIFF\alpha\sigma^2}\right)\Tr(I-PP^\top)\\
    =&\Delta^*
\end{align}
\end{subequations}
% This means that introducing noise-level dependence into the diffusion model would lead to lower loss compared to the time-shared model.
% Note that the same results for noise-level dependent model can be derived by taking the limit for the interval $\mathcal{I}$ in Eq.~\ref{eq:loss} to a single point.
This inequality formally proves that allowing the linear denoiser to be noise-level-dependent attains a minimum training loss no greater than that of the time-shared model at the same fixed $k$.
Intuitively, a time-shared model is constrained to find a single global compromise matrix $W$ across all noise levels, whereas the noise-level-dependent formulation ($W(\alpha)$) tailors its capacity to every individual noise scale $\alpha$, effectively eliminating the sub-optimality introduced by temporal averaging.

\section{Implementation Details}
% For latent diffusion, we mainly follow the setting from LightningDiT~\cite{yao2025reconstruction}, except for sampling where we use Heun solver with 50 sampling steps (NFE=99) and a classifier-free guidance (CFG)~\cite{ho2022classifier} scale of 1.5.
For latent diffusion, we adopt the training configuration of LightningDiT~\cite{yao2025reconstruction}, with the exception of the sampling procedure. We utilize the Heun solver with 50 sampling steps (NFE=99) and a Classifier-Free Guidance (CFG)~\cite{ho2022classifier} scale of 1.5.
% We find for the original flow matching setting used in LightningDiT trained 64 epochs, the performance is comparable and even better.
Notably, we observe that our method, when applied to the original flow matching setting of LightningDiT, achieves comparable or superior performance for 64 training epochs.

% For pixel diffusion, we follow the setting from JiT~\cite{li2025back} for both training and sampling, i.e., Heun solver with 50 sample steps and a CFG scale of 2.9.
For pixel-space diffusion, we strictly follow the experimental protocol from JiT~\cite{li2025back} for both training and inference, employing the Heun solver with 50 steps and a CFG scale of 2.9.

% All settings are summarized in Table~\ref{tab:config}.
A comprehensive summary of all architectural and sampling hyperparameters is provided in Table~\ref{tab:config}.
Each experiment is conducted with 8 RTX PRO 6000 GPUs.

\begin{table}[!htbp]
\centering
\caption{\textbf{Configurations of experiments.}}
% \vspace{28em}
\label{tab:config}
\vspace{1.5pt}
\resizebox{1.0\width}{!}{
\tablestyle{4pt}{1.02}
\begin{tabular}{l | x{96} x{96}}
 & \textbf{Latent-space Diffusion} & \textbf{Pixel-space Diffusion} \\
\shline
 & \textbf{LightningDiT-XL} & \textbf{JiT-B} \\
\shline
\rowcolor[gray]{0.9}\multicolumn{3}{l}{\textbf{architecture}} \\
tokenizer & VA-VAE~\cite{yao2025reconstruction} & - \\
depth & 28 & 12 \\
hidden dim & 1152 & 768 \\
heads & 16 & 12 \\
image size & 256 & 256 (other settings: 512) \\
patch size & 1 & \texttt{image\_size} / 16 \\
bottleneck & - & 128 \\
dropout & 0 & 0 \\
in-context class tokens & - & 32 \\
in-context start block & - & 4 \\
\midline
\rowcolor[gray]{0.9}\multicolumn{3}{l}{\textbf{training}} \\
loss & \multicolumn{2}{c}{\vv-loss} \\
epochs & 64 (ablation), 800 & 200 (ablation), 600 \\
warmup epochs \cite{Goyal2017} & 0 & 5 \\
optimizer & \multicolumn{2}{c}{Adam \cite{adam2014method}, $\beta_1, \beta_2=0.9, 0.95$} \\
batch size & \multicolumn{2}{c}{1024} \\
learning rate & \multicolumn{2}{c}{2e-4} \\ 
learning rate schedule & \multicolumn{2}{c}{constant} \\
weight decay & \multicolumn{2}{c}{0} \\ 
ema decay & \multicolumn{2}{c}{\{0.9996, 0.9998, 0.9999\}} \\
\multirow{2}{*}{time sampler} & \multicolumn{2}{c}{$\text{logit}(t){\sim}\mathcal{N}(\mu, \sigma^2)$} \\
 & $\mu=0.0, \sigma=1.0$ & $\mu=-0.8, \sigma=0.8$ \\
noise scale & 1.0 & 1.0 $\times$ \texttt{image\_size} / 256 \\
\multirow{2}{*}{clip of $(1-t)$ in division} & 0.05 for~\xx-prediction & \multirow{2}{*}{0.05} \\
 & and 0 for others & \\
% clip of $(1-t)$ in division & \multicolumn{2}{c}{0.05} \\
class token drop (for CFG) & \multicolumn{2}{c}{0.1} \\
\midline
\rowcolor[gray]{0.9}\multicolumn{3}{l}{\textbf{sampling}} \\
ODE solver & \multicolumn{2}{c}{Heun \cite{heun1900neue}} \\
ODE steps & \multicolumn{2}{c}{50} \\
time steps & \multicolumn{2}{c}{linear in [0.0, 1.0]} \\
% CFG scale sweep range \cite{ho2022classifier} & \multicolumn{2}{c}{[1.0, 4.0]}  \\
CFG scale \cite{ho2022classifier} & 1.5 & 2.9 \\
CFG interval \cite{kynkaanniemi2024applying} & \multicolumn{2}{c}{{[0.1, 1]}} \\
\end{tabular}
}
% \vspace{-.5em}
\end{table}

\section{Broader Impacts}
By automating the selection of optimal prediction targets, our work enhances training efficiency and reduces the computational energy required to develop high-quality generative models. However, as with any advancement in generative AI, these improvements in model performance also lower the barrier for the potential misuse of diffusion models in creating harmful or deceptive content.

\newpage

\end{document}